\documentclass{article} 
\usepackage{iclr2021_conference,times}

\usepackage{amsmath,amsfonts,bm}

\def\eqref#1{equation~\ref{#1}}

\def\1{\bm{1}}

\DeclareMathAlphabet{\mathsfit}{\encodingdefault}{\sfdefault}{m}{sl}
\SetMathAlphabet{\mathsfit}{bold}{\encodingdefault}{\sfdefault}{bx}{n}

\DeclareMathOperator*{\argmin}{arg\,min}

\usepackage{hyperref}
\usepackage{natbib}
\usepackage{bm}
\usepackage{url}
\usepackage{color}
\usepackage{amsmath,amssymb,amsthm,amsfonts}
\newtheorem{Lemma}{Lemma}
\newtheorem{Theorem}[Lemma]{Theorem}

\usepackage{algorithm,algorithmic}
\usepackage{footnote}
\usepackage{graphicx}
\usepackage{subfigure}
\usepackage{epstopdf}
\usepackage{enumerate}
\usepackage{graphicx}
\usepackage[centerlast]{caption}
\usepackage{xcolor}

\title{ The Role of Momentum Parameters in the Optimal Convergence of Adaptive Polyak's Heavy-ball Methods}

\newcommand*\samethanks[1][\value{footnote}]{\footnotemark[#1]}

\author{Wei Tao \thanks{Equal contribution} \\
Institute of Evaluation and Assessment Research\\
Academy of Military Science\\
Beijing, China \\
\texttt{wtao\_plaust@163.com} \\
\And
Sheng Long \samethanks\\
Department of Information Engineering\\
Army Academy of Artillery and Air Defense \\
Hefei, China \\
\texttt{ls15186322349@163.com} \\
\And
Gaowei Wu, Qing Tao \thanks{Corresponding author} \\
Institute of Automation\\
Chinese Academy of Sciences\\ 
Beijing, China \\
\texttt{\{gaowei.wu,qing.tao\}@ia.ac.cn}

}

\iclrfinalcopy 
\begin{document}

\maketitle

\begin{abstract}
The adaptive stochastic gradient descent (SGD) with momentum has been widely adopted in deep learning as well as convex optimization. In practice, the last iterate is commonly used as the final solution to make decisions. However, the available regret analysis and the setting of constant momentum parameters only guarantee the optimal convergence of the averaged solution. In this paper, we fill this theory-practice gap by investigating the convergence of the last iterate (referred to as {\it individual convergence}), which is a more difficult task than convergence analysis of the averaged solution. Specifically, in the constrained convex cases, we prove that the adaptive Polyak's Heavy-ball (HB) method, in which only the step size is updated using the exponential moving average strategy, attains an optimal individual convergence rate of $O(\frac{1}{\sqrt{t}})$, as opposed to the optimality of $O(\frac{\log t}{\sqrt {t}})$ of SGD, where $t$ is the number of iterations. Our new analysis not only shows how the HB momentum and its time-varying weight help us to achieve the acceleration in convex optimization but also gives valuable hints how the momentum parameters should be scheduled in deep learning. Empirical results on optimizing convex functions and training deep networks validate the correctness of our convergence analysis and demonstrate the improved performance of the adaptive HB methods.
\end{abstract}

\section{Introduction}

One of the most popular optimization algorithms in deep learning is the momentum method \citep{krizhevsky2012imagenet}. The first momentum can be traced back to the pioneering work of Polyak's heavy-ball (HB) method \citep{polyak1964some}, which helps accelerate stochastic gradient descent (SGD) in the relevant direction and dampens oscillations \citep{ruder2016overview}. Recent studies also find that the HB momentum has the potential to escape from the local minimum and saddle points \citep{ochs2014ipiano,sun2019heavy}. From the perspective of theoretical analysis, HB enjoys a smaller convergence factor than SGD when the objective function is twice continuously differentiable and strongly convex \citep{ghadimi2015global}. In nonsmooth convex cases, with suitably chosen step size, HB attains an optimal convergence rate of $O(\frac{1}{\sqrt{t}})$ in terms of the averaged output \citep{yang2016unified}, where $t$ is the number of iterations.

To overcome the data-independent limitation of predetermined step size rules, some adaptive gradient methods have been proposed to exploit the geometry of historical data. The first algorithm in this line is AdaGrad \citep{duchi2011adaptive}. The intuition behind AdaGrad is that the seldom-updated weights should be updated with a larger step size than the frequently-updated weights. Typically, AdaGrad rescales each coordinate and estimates the predetermined step size by a sum of squared past gradient values. As a result, AdaGrad has the same convergence rate as vanilla SGD but enjoying a smaller factor especially in sparse learning problems. The detailed analysis of AdaGrad \citep{mukkamala2017variants} implies that one can derive similar convergence rates for the adaptive variants of the predetermined step size methods without additional difficulties.

Unfortunately, experimental results illustrate that AdaGrad under-performed when applied to training deep neural newtworks \citep{wilson2017marginal}. Practical experience has led to the development of adaptive methods that is able to emphasize the more recent gradients. Specifically, an exponential moving average (EMA) strategy was proposed in RMSProp to replace the cumulative sum operation \citep{tieleman2012lecture}. Adam \citep{kingma2014adam}, which remains one of the most popular optimization algorithms in deep learning till today, built upon RMSProp together with updating the search directions via the HB momentum. Generally speaking, the gradient-based momentum algorithms that simultaneously update the search directions and learning rates using the past gradients are referred to as the Adam-type methods \citep{chen2018convergence}. This kinds of methods have achieved several state-of-the-art results on various learning tasks \citep{sutskever2013importance}.

Compared with HB and AdaGrad, the main novelty of Adam lies in applying EMA to gradient estimate (first-order) and to element-wise square-of-gradients (second-order), with parameters $\beta_{1t}$ and $\beta_{2t}$ \citep{alacaoglu2020new}. However, the use of EMA causes a lot of complexities to the convergence analysis. For example, in the online setting, \citep{kingma2014adam} offered a proof that Adam would converge to the optimum. Despite its remarkable practicality, Adam suffers the non-convergence issue . To overcome its advantages, several variants such as AMSGrad and AdamNC were proposed \citep{reddi2019convergence}. Unfortunately, the best regret bound of AMSGrad is only $O(\sqrt{\log t}\sqrt {t})$ for nonsmooth convex problems \citep{reddi2019convergence}, as opposed to the optimality of $O(\sqrt {t})$ of SGD. On the other hand, EMA uses only the current gradient in updating the search direction while the original HB can use the past gradients information \citep{zou2018weighted}. This will lead the update to stagnate when $\beta_{1t}$ is very close to 1. Fortunately, such a dilemma will not appear in Polyak's HB method and a simple proof on the convergence of this kind of Adams in smooth cases has been provided \citep{Dfossez2020OnTC}.

In this paper, we will focus on the adaptive Polyak's HB method, in which only the step size is updated using EMA. Despite various practical performance report for the Adam-type methods, there still exist some gaps between theoretical guarantees and empirical success.
\begin{itemize}
\item First of all, some important regret bounds have been established to guarantee the performance of online Adam-type algorithms. Nevertheless, the online-to-batch conversion can inevitably lead the solution of the induced stochastic algorithm to take the form of averaging of all the past iterates. In practice, the last iterate is popularly used as the final solution, which has the advantage of readily enforcing the learning structure \citep{chen2012optimal}. For SGD, the convergence of the last iterate, which is referred to as {\it individual convergence} in \citep{tao2020primal}, was posed as an open problem \citep{shamir2012open}. Only recently, its optimal individual convergence rate is proved to be $O(\frac{\log t}{\sqrt{t}})$ and $O(\frac{\log t}{t})$ for general and strongly convex problems respectively \citep{harvey2019tight,jain2019making}. Despite enjoying the optimal averaging convergence \citep{yang2016unified}, as far as we know, the individual convergence has not been discussed for the adaptive HB.
\item Secondly, the momentum technique is often claimed as an accelerated strategy in machine learning community. However, almost all the theoretical analysis is only limited to the Nesterov's accelerated gradient (NAG) \citep{nesterov27method} method especially in smooth cases \citep{hu2009accelerated,liu2018accelerating}, which accelerates the rate of SGD from $O(\frac{1}{t})$ to $O(\frac{1}{t^2})$. While the individual convergence of HB is also concerned in some papers  \citep{Sebbouh2020OnTC,Sun2019NonErgodicCA}, the considered problem is limited to smooth and the derived rate is not optimal in convex cases. It is discovered that NAG is capable of accelerating the rate of individual convergence of SGD from $O(\frac{\log t}{\sqrt{t}})$ to $O(\frac{1}{\sqrt t})$ \citep{srsg2019} in nonsmooth convex cases. Nevertheless, there is still a lack of the report about the acceleration of the adaptive HB.
\item Finally, in practice, almost all the momentum and Adam-type algorithms are often used with a constant momentum parameter $\beta_{1t}$ (typically between 0.9 and 0.99). In theory, regret guarantees in the online Adam require a rapidly increasing $\beta_{1t}\rightarrow 1$ schedule, which is also considered in \citep{sutskever2013importance,Orvieto2019TheRO}. This gap is recently bridged by getting the same regret bounds as that in \citep{reddi2019convergence} with a constant $\beta_{1t}$. In each state-of-the-art deep learning library (e.g. TensorFlow, PyTorch and Keras), HB is named as SGD with momentum and $\beta_{1t}$ is empirically set to 0.9 \citep{ruder2016overview}. Despite its intuition in controlling the number of  forgotten past gradients and its guarantee in optimal averaging convergence \citep{yang2016unified}, how $\beta_{1t}$ affects the optimal individual convergence has not been discussed \citep{gitman2019understanding}.
 \end{itemize}

The goal of this paper is to close a theory-practice gap when using HB to train the deep neural networks as well as optimize the convex objective functions. Specifically,

\begin{itemize}
\item By setting $\beta_{1t}=\frac{t}{t+2}$, we prove that the adaptive HB attains an optimal individual convergence rate of $O(\frac{1}{\sqrt{t}})$ (Theorem \ref{adaptive individual bound}), as opposed to the optimality of $O(\frac{\log t}{\sqrt {t}})$ of SGD. Our proof is different from all the existing analysis of averaging convergence. It not only provides a theoretical guarantee for the acceleration of HB but also clarifies how the momentum and its parameter $\beta_{1t}$ help us to achieve the optimal individual convergence.
\item If $0\leq \beta_{1t} \equiv \beta < 1$, we prove that the adaptive HB attains optimal averaging convergence (Theorem \ref{adaptive averaging bound}). To guarantee the optimal individual convergence, Theorem \ref{adaptive individual bound} suggests that time-varying $\beta_{1t}$ can be adopted. Note $\beta_{1t}=\frac{t}{t+2} \rightarrow 1$, thus our new convergence analysis not only offers an interesting explanation why we usually restrict $\beta_{1t}\rightarrow 1$ but also gives valuable hints how the momentum parameters should be scheduled in deep learning.
\end{itemize}

We mainly focus on the proof of optimal individual convergence of HB (Theorem \ref{individual bound}, Appendix A). The analysis of averaging convergence (Theorem \ref{averaging bound}) is simpler. Their extensions to include adaptive strategy are slightly more complex (Theorem \ref{adaptive individual bound} and \ref{adaptive averaging bound}), but it is similar to the proof of AdaGrad \citep{mukkamala2017variants} and the details can be found in supplementary material.

\section{Problem Statement and Related Work}

Consider the following optimization problem,
\begin{equation}\label{constrained minimization}
\min f(\mathbf{w}), \ s. t. \  \mathbf{w} \in \mathbf{Q}.
\end{equation}
where $\mathbf{Q} \subseteq \mathbb{R}^{d}$ is a closed convex set and $f(\mathbf{w})$ is a convex function. Denote that $\mathbf{w}^{*}$ is an optimal solution and $P$ is the projection operator on $ \mathbf{Q} $. Generally, the averaging convergence is defined as
\begin{equation}\label{averaged convergence}
f(\mathbf{\bar w}_{t}) - f(\mathbf{w}^{*}) \leq \epsilon(t),
\end{equation}
where $\mathbf{\bar w}_{t} = \sum_{i=1}^{t} \mathbf{w}_{i} $ and $\epsilon(t)$ is the convergence bound about \textit{t}. By contrast, the \textit{individual convergence} is described as
\begin{equation}\label{individual convergence}
f(\mathbf{w}_{t}) - f(\mathbf{w}^{*}) \leq \epsilon(t).
\end{equation}
Throughout this paper, we use $\mathbf{g}(\mathbf{w}_t)$ to denote the subgradient of $f$ at $\mathbf{w}_t$. Projected subgradient descent (PSG) is one of the most fundamental algorithms for solving problem (\ref{constrained minimization}) \citep{bertsekas2003convex}, and the iteration of which is
$$
\mathbf{w}_{t+1} = P[\mathbf{w}_{t} - \alpha_t  \mathbf{g}(\mathbf{w}_t)],
$$
where $\alpha_t>0$ is the step size. To analyze the convergence, we need the following assumption.

{\bf Assumption 1}.  {\it Assume that there exists a number $M>0$ such that
$$\|\mathbf{g}(\mathbf{w})\| \leq M, \ \forall \mathbf{w} \in \mathbf{Q}. $$}
It is known that the optimal bound for the nonsmooth convex  problem (\ref{constrained minimization}) is $O(\frac{1}{\sqrt{t}})$ \citep{nemirovsky1983problem}. PSG can attain this optimal convergence rate in terms of the averaged output while its optimal individual rate is only $O(\frac{\log t}{\sqrt {t}})$ \citep{harvey2019tight,jain2019making}.

When $\mathbf{Q} = \mathbb{R}^{N}$, the regular HB for solving the unconstrained problem (\ref{constrained minimization}) is
\begin{equation}\label{HB}
\mathbf{w}_{t+1} = \mathbf{w}_{t} - \alpha_t  \mathbf{g}(\mathbf{w}_{t})+\beta_{t}(\mathbf{w}_{t}-\mathbf{w}_{t-1}).
\end{equation}
If $0\leq \beta_{t} \equiv \beta < 1$, the key property of HB is that it can be reformulated as \citep{ghadimi2015global}
\begin{equation}\label{equivalent HB}
\mathbf{w}_{t+1}+\mathbf{p}_{t+1}=\mathbf{w}_{t}+\mathbf{p}_{t}-\displaystyle\frac{\alpha_t}{1-\beta}\mathbf{g}(\mathbf{w}_{t}),\text{where}\ \mathbf{p}_t=\frac{\beta}{1-\beta}(\mathbf{w}_{t}-\mathbf{w}_{t-1}).
\end{equation}
Thus its convergence analysis makes almost no difference to that of PSG. Especially, if $\alpha_t\equiv \frac{\alpha}{\sqrt{T}}$, its averaging convergence rate is $O(\frac{1}{\sqrt{T}})$  \citep{yang2016unified}, where $T$ is the total number of iterations.

Simply speaking, the regular Adam \citep{kingma2014adam} takes the form of
$$
 \mathbf{w}_{t+1}=\mathbf{w}_{t}-\displaystyle\frac{\alpha}{\sqrt{t}}V_{t}^{-\frac{1}{2}}\hat{\mathbf{g}}_{t},
$$
where $\hat{\mathbf{g}}(\mathbf{w}_t)$ is a unbiased estimation of $ \mathbf{g}(\mathbf{w}_t)$, and
$$
 \hat{\mathbf{g}}_{t}=\beta_{1t}\hat{\mathbf{g}}_{t-1}+(1-\beta_{1t})\hat{\mathbf{g}}(\mathbf{w}_t), \ V_{t}=\beta_{2t}V_{t-1}+(1-\beta_{2t})\text{diag}(\hat{\mathbf{g}}(\mathbf{w}_t) \hat{\mathbf{g}}(\mathbf{w}_t)^{\top}).
$$

\section{Individual Convergence of HB}

To solve the constrained problem (\ref{constrained minimization}), HB can be naturally reformulated as
\begin{equation}\label{constrained HB}
\mathbf{w}_{t+1} = P_{\mathbf{Q}}[\mathbf{w}_{t} - \alpha_t  \mathbf{g}(\mathbf{w}_{t})+\beta_t(\mathbf{w}_{t}-\mathbf{w}_{t-1})].
\end{equation}
We first prove a key lemma, which extends (\ref{equivalent HB}) to the constrained and time-varying cases.

\begin{Lemma}\label{projection inequality}
\citep{bertsekas2003convex} For $ \mathbf{w} \in \mathbb{R}^{d},\mathbf{w}_{0}\in \mathbf{Q}$,
$$\langle \mathbf{w}-\mathbf{w}_{0}, \mathbf{u}-\mathbf{w}_{0}\rangle\leq 0$$
for all $\mathbf{u}\in \mathbf{Q}$ if and only if $\mathbf{w}_{0}=P(\mathbf{w})$.
\end{Lemma}

\begin{Lemma}\label{constrained and unconstrained}
Let $\{{\mathbf{w}}_{t}\}_{t=1}^{\infty}$ be generated by HB (\ref{constrained HB}). Let
$$
\mathbf{p}_t=t(\mathbf{w}_{t}-\mathbf{w}_{t-1}),\ \beta_t=\displaystyle\frac{t}{t+2},\ \alpha_t=\displaystyle\frac{\alpha}{(t+2)\sqrt{t}}.
$$
Then HB (\ref{constrained HB}) can be reformulated as
\begin{equation}\label{new-finding}
\mathbf{w}_{t+1}+\mathbf{p}_{t+1}=P_{\mathbf{Q}}[\mathbf{w}_{t}+\mathbf{p}_{t}-\frac{\alpha}{\sqrt t}\mathbf{g}(\mathbf{w}_t)].
\end{equation}
\end{Lemma}

{\bf Proof}.  The projection operation can be rewritten as an optimization problem \citep{duchi2018introductory}, i.e., $\mathbf{w}_{t+1} = P_{\mathbf{Q}}[\mathbf{w}_{t} - \alpha_t  \mathbf{g}(\mathbf{w}_t)+\beta_t(\mathbf{w}_{t}-\mathbf{w}_{t-1})]$ is equivalent to
\begin{equation}\label{HB-regularization}
 \mathbf{w}_{t+1} = \arg\min_{\mathbf{w} \in \mathbf{Q}}\{ \alpha_t \langle \mathbf{g}(\mathbf{w}_t),\mathbf{w}\rangle
 +\frac{1}{2} \|\mathbf{w}-\mathbf{w}_{t}-\beta_t(\mathbf{w}_{t}-\mathbf{w}_{t-1})\|^2 \}.
\end{equation}
Then, $\forall\mathbf{w}\in \mathbf{Q}$, we have
$$
\langle\mathbf{w}_{t+1}-\mathbf{w}_{t}-\beta_t(\mathbf{w}_{t}-\mathbf{w}_{t-1})+ \alpha_t  \mathbf{g}(\mathbf{w}_t),\mathbf{w}_{t+1}-\mathbf{w}\rangle \leq 0.
$$
This is
\begin{equation}\label{proj1}
\langle\mathbf{w}_{t+1}+\mathbf{p}_{t+1}-(\mathbf{w}_{t}+\mathbf{p}_{t})+\frac{\alpha}{\sqrt t}\mathbf{g}(\mathbf{w}_t),\mathbf{w}_{t+1}-\mathbf{w}\rangle \leq 0
\end{equation}
Specifically,
\begin{equation}\label{proj2}
\langle\mathbf{w}_{t+1}+\mathbf{p}_{t+1}-(\mathbf{w}_{t}+\mathbf{p}_{t})+\frac{\alpha}{\sqrt t}\mathbf{g}(\mathbf{w}_t),\mathbf{w}_{t+1}-\mathbf{w}_{t}\rangle \leq 0.
\end{equation}
From (\ref{proj1}) and (\ref{proj2}),
$$
\langle\mathbf{w}_{t+1}+\mathbf{p}_{t+1}-(\mathbf{w}_{t}+\mathbf{p}_{t})+\frac{\alpha}{\sqrt t}\mathbf{g}(\mathbf{w}_t),\mathbf{w}_{t+1}-\mathbf{w}+(t+1)(\mathbf{w}_{t+1}-\mathbf{w_t})\rangle \leq 0.
$$
i.e.,
$$
\langle\mathbf{w}_{t+1}+\mathbf{p}_{t+1}-(\mathbf{w}_{t}+\mathbf{p}_{t})+\frac{\alpha}{\sqrt t}\mathbf{g}(\mathbf{w}_t),\mathbf{w}_{t+1}+\mathbf{p}_{t+1}-\mathbf{w}\rangle \leq 0.
$$
Using Lemma \ref{projection inequality}, Lemma \ref{constrained and unconstrained} is proved.

Due to the non-expansive property of $P_{\mathbf{Q}}$ \citep{bertsekas2003convex}, Lemma \ref{constrained and unconstrained} implies that the convergence analysis for unconstrained problems can be applied to analyze the constrained problems.

\begin{Theorem}\label{individual bound}
Assume that $\mathbf{Q}$ is bounded. Let $\{{\mathbf{w}}_{t}\}_{t=1}^{\infty}$ be generated by HB (\ref{constrained HB}). Set
$$\beta_t=\displaystyle\frac{t}{t+2} \ \text{and} \  \alpha_t=\displaystyle\frac{\alpha}{(t+2)\sqrt{t}}.$$
Then
$$
f(\mathbf{w}_{t})-f(\mathbf{w}^*) \leq O( \frac{1}{\sqrt{t}}).
$$
\end{Theorem}

{\bf Proof}. \ According to Lemma \ref{constrained and unconstrained},
$$\|\mathbf{w}^*-(\mathbf{w}_{t+1}+\mathbf{p}_{t+1})\|^2 \leq  \|\mathbf{w}^*-(\mathbf{w}_{t}+\mathbf{p}_{t})+\frac{\alpha}{\sqrt t}\mathbf{g}(\mathbf{w}_t)\|^2. $$
$
\begin{aligned}
& \|\mathbf{w}^*-(\mathbf{w}_{t}+\mathbf{p}_{t})+\frac{\alpha}{\sqrt t}\mathbf{g}(\mathbf{w}_t)\|^2\\
& =\|\mathbf{w}^*-(\mathbf{w}_{t}+\mathbf{p}_{t})\|^2+\|\frac{\alpha}{\sqrt t}\mathbf{g}(\mathbf{w}_t)\|^2 +2\langle\frac{\alpha}{\sqrt t}\mathbf{g}(\mathbf{w}_t),\mathbf{w}^*-\mathbf{w}_{t}\rangle+2\langle\frac{\alpha t}{\sqrt t}\mathbf{g}(\mathbf{w}_t),\mathbf{w}_{t-1}-\mathbf{w}_{t}\rangle
\end{aligned}
$
Note
$$\langle\mathbf{g}(\mathbf{w}_t),\mathbf{w}^*-\mathbf{w}_{t}\rangle \leq f(\mathbf{w}^*)-f(\mathbf{w}_{t}),\ \langle\mathbf{g}(\mathbf{w}_t),\mathbf{w}_{t-1}-\mathbf{w}_{t}\rangle \leq f(\mathbf{w}_{t-1})-f(\mathbf{w}_{t}). $$
Then
$$
\begin{aligned}
& (t+1)(f(\mathbf{w}_t)-f(\mathbf{w}^*)) \\
& \leq t(f(\mathbf{w}_{t-1})-f(\mathbf{w}^*))+ \frac{\sqrt t}{2\alpha} \|\mathbf{w}^*-(\mathbf{w}_{t}+\mathbf{p}_{t})\|^2
 -\frac{\sqrt t}{2\alpha} \|\mathbf{w}^*-(\mathbf{w}_{t+1}+\mathbf{p}_{t+1})\|^2 + \frac{\alpha}{2\sqrt t}\|\mathbf{g}(\mathbf{w}_t)\|^2.
\end{aligned}
$$
Summing this inequality from $k=1$ to $t$, we obtain
$$
\begin{aligned}
& (t+1)(f(\mathbf{w}_t)-f(\mathbf{w}^*))\\
& \leq f(\mathbf{w}_{0})-f(\mathbf{w}^*)+ \sum_{k=1}^{t} {\frac{\alpha}{2\sqrt k}\|\mathbf{g}(\mathbf{w}_k)\|^2} + \sum_{k=1}^{t}\big[ \frac{\sqrt k}{2\alpha} (\|\mathbf{w}^*-(\mathbf{w}_{k}+\mathbf{p}_{k})\|^2  - \|\mathbf{w}^*-(\mathbf{w}_{k+1}+\mathbf{p}_{k+1})\|^2) \big].
\end{aligned}
$$
Note
$$\sum_{k=1}^{t} {\frac{1}{2\sqrt k}\|\mathbf{g}(\mathbf{w}_k)\|^2}\leq \sqrt t M^{2}. $$
and
$$
\begin{aligned}
&  \sum_{k=1}^{t}\big[ \frac{\sqrt k}{2} (\|\mathbf{w}^*-(\mathbf{w}_{k}+\mathbf{p}_{k})\|^2  - \|\mathbf{w}^*-(\mathbf{w}_{k+1}+\mathbf{p}_{k+1})\|^2) \big].\\
& \leq \frac{1}{2} \|\mathbf{w}^{*}-(\mathbf{w}_{1}+\mathbf{p}_{1})\|^2- \frac{\sqrt t}{2} \|\mathbf{w}^*-(\mathbf{w}_{t+1}+\mathbf{p}_{t+1})\|^2 +\sum_{k=2}^{t}(\frac{\sqrt k}{2}-\frac{\sqrt {k-1}}{2}) \|\mathbf{w}^*-(\mathbf{w}_{k}+\mathbf{p}_{k})\|^2.
\end{aligned}
$$
Since $\mathbf{Q}$ is a bounded set, there exists a positive number $M_0 >0$ such that
$$
\|\mathbf{w}^*-(\mathbf{w}_{t+1}+\mathbf{p}_{t+1})\|^2 \leq M_0, \forall t \geq 0.
$$
Therefore
$$
\begin{aligned}
& (t+1)[f(\mathbf{w}_t)-f(\mathbf{w}^*)] \leq f(\mathbf{w}_{0})-f(\mathbf{w}^*)+ \alpha\sqrt t M^2  +\frac{\sqrt t}{2\alpha} M_{0}.
\end{aligned}
$$

This completes the proof of Theorem \ref{individual bound}.

It is necessary to give some remarks about Theorem \ref{individual bound}.
\begin{itemize}
\item In nonsmooth convex cases, Theorem \ref{individual bound} shows that the individual convergence rate of SGD can be accelerated from $O(\frac{\log t}{\sqrt{t}})$ to $O(\frac{1}{\sqrt{t}})$ via the HB momentum. The proof here clarifies how the HB-type momentum $\mathbf{w}_{t}-\mathbf{w}_{t-1}$ and its time-varying weight $\beta_t$ help us to derive the optimal individual convergence.
\item The convergence analysis in Theorem \ref{individual bound} is obviously different from the regret analysis in all the available papers, this is because the connection between $f(\mathbf{w}_{t})-f(\mathbf{w}^{*})$ and $f(\mathbf{w}_{t-1})-f(\mathbf{w}^{*})$ should be established here. It can be seen that seeking an optimal individual convergence is more difficult than the analysis of averaging convergence in many papers such as \citep{zinkevich2003online} and \citep{yang2016unified}.
\item We can get a stochastic HB by replacing the subgradient $ \mathbf{g}(\mathbf{w}_t)$ in (\ref{constrained HB}) with its unbiased estimation $ \hat{\mathbf{g}}(\mathbf{w}_t)$. Such substitution will not influence our convergence analysis. This means that we can get $\mathbb{E} [f(\mathbf{w}_{t})-f(\mathbf{w}^{*})] \leq O(\frac{1}{\sqrt{t}})$ under the same assumptions.
    \end{itemize}

If $\beta_t$ remains a constant, we can get the averaging convergence rate, in which the proof of the first part is similar to Lemma \ref{constrained and unconstrained} and that of the second part is similar to online PSG \citep{zinkevich2003online}.

\begin{Theorem}\label{averaging bound}
Assume that $\mathbf{Q}$ is bounded and $0\leq \beta_{t} \equiv \beta < 1$. Let $\{{\mathbf{w}}_{t}\}_{t=1}^{\infty}$ be generated by HB (\ref{constrained HB}). Set
$$\mathbf{p}_t=\frac{\beta}{1-\beta}(\mathbf{w}_{t}-\mathbf{w}_{t-1}) \ \text{and} \  \alpha_t=\displaystyle\frac{\alpha}{\sqrt{t}}.$$
Then we have
$$\mathbf{w}_{t+1}+\mathbf{p}_{t+1}=P_{\mathbf{Q}}[\mathbf{w}_{t}+\mathbf{p}_{t}-\displaystyle\frac{\alpha_t}{1-\beta}\mathbf{g}(\mathbf{w}_{t})],\  f(\frac{1}{t}\sum_{k=1}^{t} \mathbf{w}_k)-f(\mathbf{w}^{*}) \leq O(\frac{1}{\sqrt{t}}). $$
\end{Theorem}

If $\mathbf{Q}$ is not bounded, the boundness of sequence $\|\mathbf{w}^*-(\mathbf{w}_{t+1}+\mathbf{p}_{t+1})\|$ can not be ensured, which may lead to the failure of Theorem \ref{averaging bound}. Fortunately, like that in \citep{yang2016unified},  $\mathbb{E}[f(\frac{1}{T}\sum_{k=1}^{T} \mathbf{w}_k)-f(\mathbf{w}^{*})] \leq O(\frac{1}{\sqrt{T}})$ still holds, but we need to set $\alpha_t\equiv \frac{\alpha}{\sqrt{T}}$, where $T$ is the total number of iterations.

\section{Extension to Adaptive Cases}

It is easy to find that HB (\ref{new-finding}) is in fact a gradient-based algorithm with predetermined step size $\frac{\alpha}{\sqrt t}$. Thus its adaptive variant with EMA can be naturally formulated as
\begin{equation}\label{adaptive HB}
\mathbf{w}_{t+1} = P_{\mathbf{Q}}[\mathbf{w}_{t} - \displaystyle\frac{\alpha \beta_{1t}}{t\sqrt t} V_{t}^{-\frac{1}{2}} \hat{\mathbf{g}}(\mathbf{w}_t)+ \beta_{1t}(\mathbf{w}_{t}-\mathbf{w}_{t-1})].
\end{equation}
where
$$\beta_{1t}=\displaystyle\frac{t}{t+2},\ V_{t}=\beta_{2t}V_{t-1}+(1-\beta_{2t})\text{diag}(\hat{\mathbf{g}}(\mathbf{w}_t) \hat{\mathbf{g}}(\mathbf{w}_t)^{\top}).
$$

The detailed steps of the adaptive HB are shown in Algorithm \ref{alg.hb}.

\begin{algorithm}
 \renewcommand{\algorithmicrequire}{\textbf{Input:}}
 \renewcommand{\algorithmicensure}{\textbf{Output:}}
 \caption{Adaptive HB}
  \begin{algorithmic}[1]\label{alg.hb}
  \REQUIRE momentum parameters $\beta_{1t}$, $\beta_{2t}$, constant $\delta > 0$, the total number of iterations $T$
  \STATE Initialize $\mathbf{w}_0 = \mathbf{0}$, $V_0 = \mathbf{0}_{d\times d}$
  \REPEAT
  \STATE  $\hat{\mathbf{g}}_t(\mathbf{w}_t) = \nabla f_{t}(\mathbf{w}_t)$,
        \STATE $V_t = \beta_{2t}V_{t-1}+(1-\beta_{2t})diag(\hat{\mathbf{g}}_t(\mathbf{w}_t){\hat{\mathbf{g}}_t(\mathbf{w}_t)}^\top)$,
        \STATE $\hat{V_t} = V_t^{\frac{1}{2}}+\frac{\delta }{\sqrt{t}}I_d$,
       \STATE $\mathbf{w}_{t+1} = P_{\mathbf{Q}}[\mathbf{w}_{t} - \displaystyle\frac{\alpha \beta_{1t}}{t\sqrt t} \hat{V_t}^{-1} \hat{\mathbf{g}}(\mathbf{w}_t)+ \beta_{1t}(\mathbf{w}_{t}-\mathbf{w}_{t-1})]$,
  \UNTIL{$t = T$}
  \ENSURE $\mathbf{w}_{T}$
 \end{algorithmic}
\end{algorithm}

\begin{Theorem}\label{adaptive individual bound}
Assume that $\mathbf{Q}$ is a bounded set. Let $\{{\mathbf{w}}_{t}\}_{t=1}^{\infty}$ be generated by the adaptive HB (Algorithm \ref{alg.hb}). Denote $\mathbf{p}_t=t(\mathbf{w}_{t}-\mathbf{w}_{t-1})$. Suppose that $\beta_{1t}  =  \frac{t}{t+2}$ and $ 1-\frac{1}{t} \leq \beta_{2t} \leq 1-\frac{\gamma}{t} $ for some $ 0 \textless \gamma \leq 1$. Then
\begin{equation}\label{adaptive projection}
\mathbf{w}_{t+1}+\mathbf{p}_{t+1}=P_{\mathbf{Q}}[\mathbf{w}_{t}+\mathbf{p}_{t}- \displaystyle\frac{\alpha}{\sqrt t} \hat{V_{t}}^{-1}\hat{\mathbf{g}}(\mathbf{w}_t)]
\end{equation}
$$\mathbb{E}[f(\mathbf{w}_{t})-f(\mathbf{w}^*)] \leq O( \frac{1}{\sqrt{t}}).
$$
\end{Theorem}

The proof of (\ref{adaptive projection}) is identical to that of Lemma \ref{constrained and unconstrained}. It is easy to find that (\ref{adaptive projection}) is an adaptive variant of (\ref{new-finding}). This implies that the proof of the second part is similar to that of AdaGrad \citep{mukkamala2017variants}. When $0\leq \beta_{1t} \equiv \beta < 1$, the adaptive variant of HB (\ref{constrained HB}) is
 \begin{equation}\label{adaptive HB constant}
\mathbf{w}_{t+1} = P_{\mathbf{Q}}[\mathbf{w}_{t} - \displaystyle\frac{\alpha}{\sqrt t} V_{t}^{-\frac{1}{2}} \hat{\mathbf{g}}(\mathbf{w}_t)+ \beta(\mathbf{w}_{t}-\mathbf{w}_{t-1})].
\end{equation}
where
$$\ V_{t}=\beta_{2t}V_{t-1}+(1-\beta_{2t})\text{diag}(\hat{\mathbf{g}}(\mathbf{w}_t) \hat{\mathbf{g}}(\mathbf{w}_t)^{\top}).
$$

Similar to the proof of Theorem \ref{adaptive individual bound}, we can get the following averaging convergence.
\begin{Theorem}\label{adaptive averaging bound}
Assume that $\mathbf{Q}$ is bounded and $0\leq \beta_{1t} \equiv \beta < 1$ in Algorithm \ref{alg.hb}. Let $\{{\mathbf{w}}_{t}\}_{t=1}^{\infty}$ be generated by the adaptive HB ((Algorithm \ref{alg.hb})). Suppose that $ 1-\frac{1}{t} \leq \beta_{2t} \leq 1-\frac{\gamma}{t} $ for some $ 0 \textless \gamma \leq 1$. Denote $\mathbf{p}_t=\frac{\beta}{1-\beta}(\mathbf{w}_{t}-\mathbf{w}_{t-1}) $. Then $$\mathbf{w}_{t+1}+\mathbf{p}_{t+1}=P_{\mathbf{Q}}[\mathbf{w}_{t}+\mathbf{p}_{t}-\displaystyle\frac{\alpha}{(1-\beta)\sqrt t} \hat{V_{t}}^{-1}\hat{\mathbf{g}}(\mathbf{w}_{t})],\  \mathbb{E}[f(\frac{1}{t}\sum_{k=1}^{t} \mathbf{w}_k)-f(\mathbf{w}^{*})] \leq O(\frac{1}{\sqrt{t}}). $$
\end{Theorem}

It is necessary to give some remarks about Theorem \ref{adaptive individual bound} and Theorem \ref{adaptive averaging bound}.
\begin{itemize}
\item The adaptive HB is usually used with a constant $\beta_{1t}$ in deep learning. However, according to Theorem \ref{adaptive averaging bound}, the constant $\beta_{1t}$ only guarantees the optimal data-dependent averaging convergence. The convergence property of the last iterate still remains unknown.
\item  In order to guarantee the optimal individual convergence, according to Theorem \ref{adaptive individual bound}, $\beta_{1t}$ has to be time-varying. $\beta_{1t}=\frac{t}{t+2} $ can explain why usually restrict $\beta_{1t}\rightarrow 1$ in practice. It also offers a new schedule about the selection of momentum parameters in deep learning.
\end{itemize}

\section{Experiments}

In this section, we present some empirical results to validate the correctness of our convergence analysis and demonstrate the improved performance of the adaptive HB methods. The experiments on optimizing the constrained hinge loss problem are given in Appendix B. For fair comparison, we independently repeat the experiments five times and report the averaged the results.

\subsection{Training Deep Neural Networks}

This experiment is to show the improved performance of Adaptive HB on 4-layer CNN and ResNet-18. The experiments are conducted on a sever with 2 NVIDIA 2080Ti GPUs.

Datasets: MNIST (60000 training samples, 10000 test samples), CIFAR10 (50000 training samples, 10000 test samples), and CIFAR100 (50000 training samples, 10000 test samples).

Algorithms: Adam ($\alpha$, $\beta_{1t}\equiv 0.9$, $\beta_{2t}\equiv 0.999$, $\epsilon=10^{-8}$) \citep{kingma2014adam}, SGD ($\alpha_t \equiv \alpha$), SGD-momentum ($\alpha_t \equiv \alpha$, $\beta_t\equiv0.9$), AdaGrad ($\alpha_t \equiv \alpha$) \citep{duchi2011adaptive}, RMSprop ($\alpha_t \equiv \alpha$, $\beta_{2t}\equiv 0.9$, $\epsilon=10^{-8}$) \citep{tieleman2012lecture}. For our adaptive HB, $\gamma=0.1$ and $\delta=10^{-8}$. Different from the existing methods, we set $\beta_{1t} = \frac{t}{t+2}$ and $\beta_{2t} = 1-\frac{\gamma}{t}$ in Algorithm \ref{alg.hb}, where $t$ is the number of the epoch. Within each epoch, $\beta_{1t} $ and $\beta_{2t} $ remain unchanged.

Note that all methods have only one adjustable parameter $\alpha$, we choose $\alpha$ from the set of \{0.1, 0.01, 0.001, 0.0001\} for all experiments. Following \citep{mukkamala2017variants} and \citep{wang2019sadam}, we design a simple 4-layer CNN architecture that consists two convolutional layers (32 filters of size 3 $\times$ 3), one max-pooling layer (2 $\times$ 2 window and 0.25 dropout) and one fully connected layer (128 hidden units and 0.5 dropout). We also use weight decay and batch normalization to reduce over-fitting. The optimal rate is always chosen for each algorithm separately so that one achieves either best training objective or best test performance after a fixed number of epochs.

\begin{figure}[ht]
    \centering
    \includegraphics[width=\textwidth]{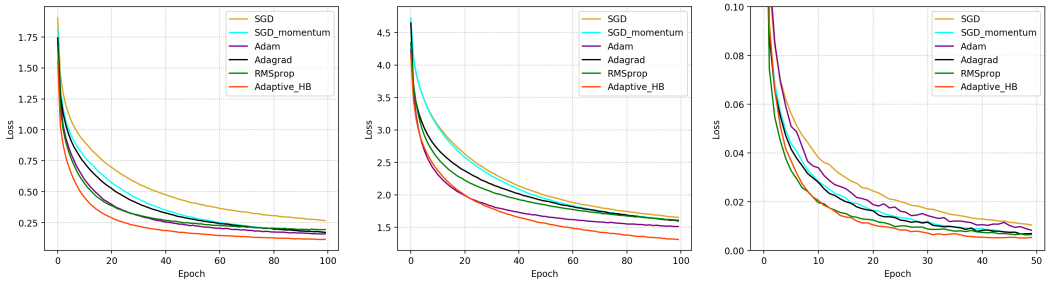}
    \caption{Training loss v.s. number of epochs on different datasets for 4-layer CNN:\\ (left) CIFAR10, (middle) CIFAR100, (right) MNIST }
    \label{fig:trainloss}
\end{figure}

\begin{figure}[ht]
    \centering
    \includegraphics[width=\textwidth]{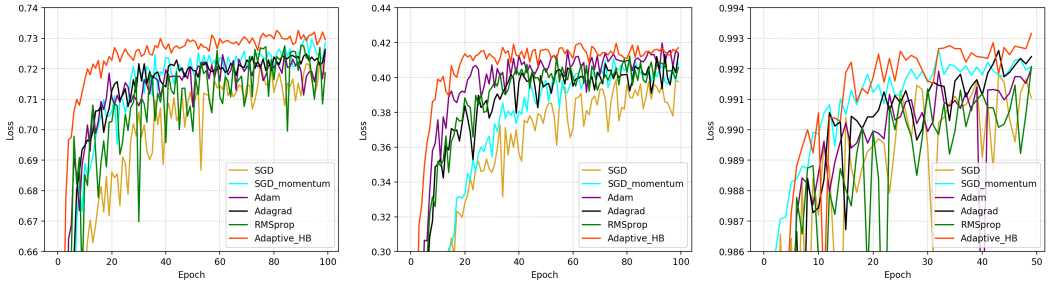}
    \caption{Test accuracy v.s. number of epochs on different datasets for 4-layer CNN:\\ (left) CIFAR10, (middle) CIFAR100, (right) MNIST }
    \label{fig:testacc}
\end{figure}

\begin{figure}[ht]
    \centering
    \includegraphics[width=\textwidth]{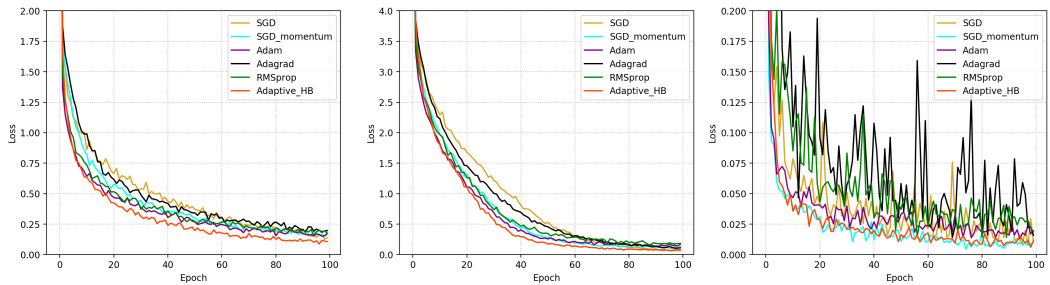}
    \caption{Training loss v.s. number of epochs on different datasets for ResNet-18:\\ (left) CIFAR10, (middle) CIFAR100, (right) MNIST}
    \label{fig:resnet trainloss}
\end{figure}

\begin{figure}[ht]
    \centering
    \includegraphics[width=\textwidth]{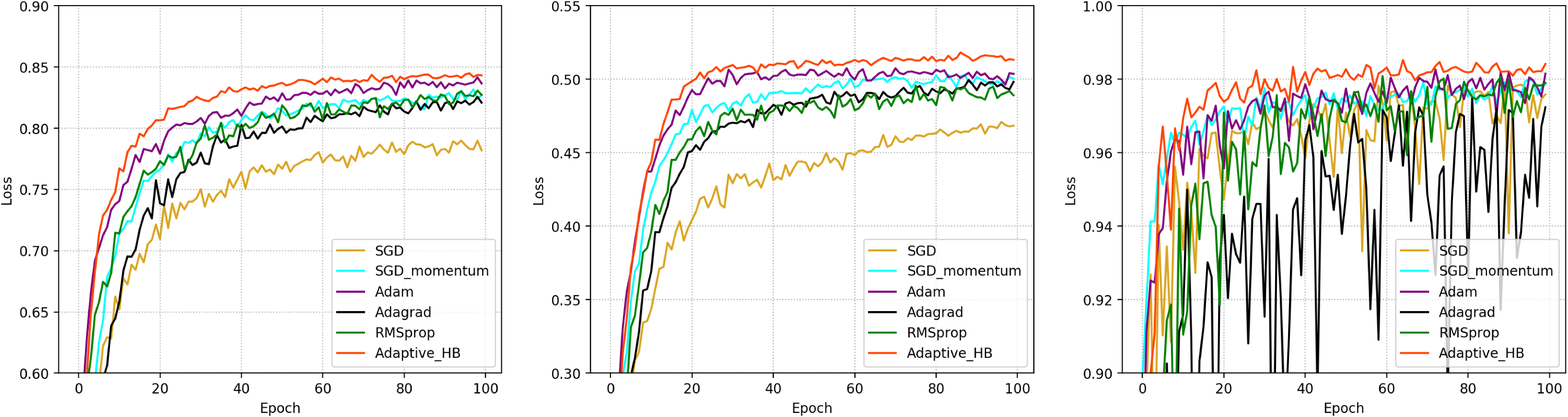}
    \caption{Test accuracy v.s. number of epochs on different datasets for ResNet-18:\\ (left) CIFAR10, (middle) CIFAR100, (right) MNIST }
    \label{fig:resnet testacc}
\end{figure}

The loss function is the cross-entropy. The training loss results are illustrated in Figure \ref{fig:trainloss} and \ref{fig:resnet trainloss}, and the test accuracy results are presented in Figure \ref{fig:testacc} and \ref{fig:resnet testacc}. As can be seen, the adaptive HB achieves the improved training loss. Moreover, this improvement also leads to good performance on test accuracy. The experimental results show that although the last iterate of the adaptive HB is optimal for general convex functions, its schedule about the momentum parameters could gain improved practical performance even in deep learning tasks.

\section{Conclusion}
In this paper, we prove that the adaptive HB method attains an optimal data-dependent individual convergence rate in the constrained convex cases, which bridges a theory-practice gap in using momentum methods to train the deep neural networks as well as optimize the convex functions. Our new analysis not only clarifies how the HB momentum and its time-varying weight $\beta_{1t}=\frac{t}{t+2}$ help us to achieve the acceleration but also gives valuable hints how its momentum parameters should be scheduled in deep learning. Empirical results on optimizing convex functions validate the correctness of our convergence analysis and several typical deep learning experiments demonstrate the improved performance of the adaptive HB.

\appendix

\bibliography{iclr2021_conference}
\bibliographystyle{iclr2021_conference}

\newpage
\section{Supplementary Material}

\subsection{Proof for Theorem \ref{averaging bound}}

Let $\{{\mathbf{w}}_{t}\}_{t=1}^{\infty}$ be generated by HB (\ref{constrained HB}) . Set
$$\mathbf{p}_t=\frac{\beta}{1-\beta}(\mathbf{w}_{t}-\mathbf{w}_{t-1}) \ \text{and} \  \alpha_t=\displaystyle\frac{\alpha}{\sqrt{t}}.$$
Then, $\forall\mathbf{w}\in \mathbf{Q}$, according to Lemma \ref{projection inequality}, we have
$$
\langle\mathbf{w}_{t+1}-\mathbf{w}_{t}-\beta(\mathbf{w}_{t}-\mathbf{w}_{t-1})+ \alpha_t  \mathbf{g}(\mathbf{w}_t),\mathbf{w}_{t+1}-\mathbf{w}\rangle \leq 0.
$$
This is
$$
\langle\frac{1}{1-\beta}(\mathbf{w}_{t+1}-\mathbf{w}_{t})-\mathbf{p}_{t}+\frac{\alpha_t}{1-\beta}\mathbf{g}(\mathbf{w}_t),
\mathbf{w}_{t+1}-\mathbf{w}\rangle \leq 0.
$$
i.e.,
\begin{equation}\label{proj3}
\langle\mathbf{w}_{t+1}+\mathbf{p}_{t+1}-(\mathbf{w}_{t}+\mathbf{p}_{t})+\frac{\alpha_t}{1-\beta}\mathbf{g}(\mathbf{w}_t),\mathbf{w}_{t+1}-\mathbf{w}\rangle \leq 0
\end{equation}
Specifically,
\begin{equation}\label{proj4}
\langle\mathbf{w}_{t+1}+\mathbf{p}_{t+1}-(\mathbf{w}_{t}+\mathbf{p}_{t})+\frac{\alpha_t}{1-\beta}\mathbf{g}(\mathbf{w}_t),
\frac{\beta(\mathbf{w}_{t+1}-\mathbf{w}_t)}{1-\beta}\rangle \leq 0
\end{equation}
From (\ref{proj3}) and (\ref{proj4}),
$$
\langle\mathbf{w}_{t+1}+\mathbf{p}_{t+1}-(\mathbf{w}_{t}+\mathbf{p}_{t})+\frac{\alpha_t}{1-\beta}\mathbf{g}(\mathbf{w}_t),
\mathbf{w}_{t+1}+\mathbf{p}_{t+1}-\mathbf{w}\rangle \leq 0.
$$
Using Lemma \ref{projection inequality}, we have
$$
\mathbf{w}_{t+1}+\mathbf{p}_{t+1}=P_{\mathbf{Q}}[\mathbf{w}_{t}+\mathbf{p}_{t}-\displaystyle\frac{\alpha_t}{1-\beta}\mathbf{g}(\mathbf{w}_{t})].
$$

Then

\begin{equation*}
\begin{split}
&\|\mathbf{w}^*-(\mathbf{w}_{t+1}+\mathbf{p}_{t+1})\|^2\\
\leq&\|\mathbf{w}^*-(\mathbf{w}_{t}+\mathbf{p}_{t})+\frac{\alpha_t}{1-\beta}\mathbf{g}(\mathbf{w}_t)\|^2\\ =&\|\mathbf{w}^*-(\mathbf{w}_{t}+\mathbf{p}_{t})\|^2+\|\frac{\alpha_t}{1-\beta}\mathbf{g}(\mathbf{w}_t)\|^2+2\langle\frac{\alpha_t}{1-\beta}\mathbf{g}(\mathbf{w}_t),\mathbf{w}^*-\mathbf{w}_{t}\rangle\\
+&2\langle\frac{\alpha_t\beta}{(1-\beta)^2}\mathbf{g}(\mathbf{w}_t),\mathbf{w}_{t-1}-\mathbf{w}_{t}\rangle
\end{split}
\end{equation*}

Note
$$\langle\mathbf{g}(\mathbf{w}_t),\mathbf{w}^*-\mathbf{w}_{t}\rangle \leq f(\mathbf{w}^*)-f(\mathbf{w}_{t}),\ \langle\mathbf{g}(\mathbf{w}_t),\mathbf{w}_{t-1}-\mathbf{w}_{t}\rangle \leq f(\mathbf{w}_{t-1})-f(\mathbf{w}_{t}). $$
Then
\begin{equation*}
\begin{split}
& \|\mathbf{w}^*-(\mathbf{w}_{t+1}+\mathbf{p}_{t+1})\|^2 \\
\leq& \|\mathbf{w}^*-(\mathbf{w}_{t}+\mathbf{p}_{t})\|^2+
\frac{\alpha_t^2}{(1-\beta)^2}\|\mathbf{g}(\mathbf{w}_t)\|^2\\
+&\frac{2\alpha_t}{1-\beta}[f(\mathbf{w}^*)-f(\mathbf{w}_{t})]+\frac{2\alpha_t\beta}{(1-\beta)^2}[f(\mathbf{w}_{t-1})-f(\mathbf{w}_{t})].
\end{split}
\end{equation*}

Rearrange the inequality, we have
\begin{equation*}
\begin{split}
\frac{2\alpha_t}{1-\beta}[f(\mathbf{w}_t)-f(\mathbf{w}^*)]\leq&\frac{2\alpha_t\beta}{(1-\beta)^2}[f(\mathbf{w}_{t-1})-f(\mathbf{w}_t)]+
\|\mathbf{w}^*-(\mathbf{w}_t+\mathbf{p}_t)\|^2\\
-&\|\mathbf{w}^*-(\mathbf{w}_{t+1}+\mathbf{p}_{t+1})\|^2+
\frac{\alpha_t^2}{(1-\beta)^2}\|\mathbf{g}(\mathbf{w}_t)\|^2.
\end{split}
\end{equation*}
i.e.,
\begin{equation*}
\begin{split}
f(\mathbf{w}_t)-f(\mathbf{w}^*)&\leq\frac{\beta}{1-\beta}[f(\mathbf{w}_{t-1})-f(\mathbf{w}_t)]+
\frac{1-\beta}{2\alpha_t}[\|\mathbf{w}^*-(\mathbf{w}_t+\mathbf{p}_t)\|^2\\
&-\|\mathbf{w}^*-(\mathbf{w}_{t+1}+\mathbf{p}_{t+1})\|^2]+
\frac{\alpha_t}{2(1-\beta)}\|\mathbf{g}(\mathbf{w}_t)\|^2.
\end{split}
\end{equation*}
Summing this inequality from $k=1$ to $t$, we obtain
$$
\begin{aligned}
& \sum_{k=1}^{t}[{f(\mathbf{w}_k)-f(\mathbf{w}^*)}]\\
\leq&\frac{\beta}{1-\beta}[f(\mathbf{w}_{0})-f(\mathbf{w}_t)]+
\frac{1-\beta}{2\alpha_1}\|\mathbf{w}^*-(\mathbf{w}_{1}+\mathbf{p}_{1})\|^2\\
-&\frac{1-\beta}{2\alpha_t}\|\mathbf{w}^*-(\mathbf{w}_{t+1}+\mathbf{p}_{t+1})\|^2
+\sum_{k=1}^{t} {\frac{\alpha_k}{2(1-\beta)}\|\mathbf{g}(\mathbf{w}_k)\|^2}\\
+&\sum_{k=2}^{t}{\|\mathbf{w}^*-(\mathbf{w}_{k}+\mathbf{p}_{k})\|^2(\frac{1-\beta}{2\alpha_k}-\frac{1-\beta}{2\alpha_{k-1}})}.
\end{aligned}
$$
i.e.,

\begin{equation}\label{proj5}
\begin{aligned}
& \sum_{k=1}^{t}[{f(\mathbf{w}_k)-f(\mathbf{w}^*)}]\\
\leq&\frac{\beta}{1-\beta}[f(\mathbf{w}_{0})-f(\mathbf{w}_t)]+
\frac{1-\beta}{2\alpha}\|\mathbf{w}^*-(\mathbf{w}_{1}+\mathbf{p}_{1})\|^2\\
+&\sum_{k=2}^{t}{\|\mathbf{w}^*-(\mathbf{w}_{k}+\mathbf{p}_{k})\|^2(\frac{(1-\beta)\sqrt{k}}{2\alpha}-\frac{(1-\beta)\sqrt{k-1}}{2\alpha})}\\
+&\sum_{k=1}^{t} {\frac{\alpha}{2(1-\beta)\sqrt{k}}\|\mathbf{g}(\mathbf{w}_k)\|^2}.
\end{aligned}
\end{equation}
Note
\begin{equation}\label{proj6}
\sum_{k=1}^{t} {\frac{1}{2\sqrt k}\|\mathbf{g}(\mathbf{w}_k)\|^2}\leq \sqrt t M^{2}.
\end{equation}
and since $\mathbf{Q}$ is a bounded set, there exists a positive number $M_0 >0$ such that
\begin{equation}\label{proj7}
\|\mathbf{w}^*-(\mathbf{w}_{t+1}+\mathbf{p}_{t+1})\|^2 \leq M_0, \forall t \geq 0.
\end{equation}
From (\ref{proj5})(\ref{proj6})(\ref{proj7}) we have,
$$
\begin{aligned}
\sum_{k=1}^{t}[{f(\mathbf{w}_k)-f(\mathbf{w}^*)}]
\leq\frac{\beta}{1-\beta}[f(\mathbf{w}_{0})-f(\mathbf{w}_t)]+
\frac{(1-\beta)\sqrt{t}M_0}{2\alpha}+\frac{\alpha\sqrt{t}M^2}{1-\beta}.
\end{aligned}
$$
By convexity of $f(\mathbf{w})$, we obtain
$$
f(\frac{1}{t}\sum_{k=1}^{t}\mathbf{w}_{k})-f(\mathbf{w}^*)\leq\frac{\beta}{(\1-\beta)t}[f(\mathbf{w}_{0})-f(\mathbf{w}_t)]+
\frac{(1-\beta)M_0}{2\alpha\sqrt{t}}+\frac{\alpha M^2}{(1-\beta)\sqrt{t}}.
$$

This completes the proof of Theorem \ref{averaging bound}.

\subsection{Proof for Theorem \ref{adaptive individual bound}}

Notation. For a positive definite matrix $H \in \Bbb R^{d\times d}$, the weighted $\ell_2$-norm is defined by $\|\mathbf{x}\|^2_H=\mathbf{x}^{\top}H\mathbf{x}$. The $H$-weighted projection $P^H_\mathbf{Q}(\mathbf{x})$ of $\mathbf{x}$ onto $\mathbf{Q}$ is defined by $P^H_\mathbf{Q}(\mathbf{x})=\mathop{\argmin}_{\mathbf{y}\in\mathbf{Q}}\|\mathbf{y}-\mathbf{x}\|^2_H$. We use $ \mathbf{g}(\mathbf{w}_k)$ to denote the subgradient of $f_k(\cdot)$ at $\mathbf{w}_k$. For the diagonal matrix sequence $\{M_k\}^t_{k=1}$, we use $m_{k,i}$ to denote the $i$-th element in the diagonal of $M_k$. We use $g_{k,i}$ to denote the $i$-th element of $\mathbf{g}(\mathbf{w}_k)$.

\begin{Lemma}\label{square gradient}
\citep{mukkamala2017variants} Suppose that $ 1-\frac{1}{t} \leq \beta_{2t} \leq 1-\frac{\gamma}{t} $ for some $ 0 \textless \gamma \leq 1$, and $ t\geq 1 $, then
$$\sum_{i=1}^{d}\sum_{k=1}^{t} {\frac{g_{k,i}^2}{\sqrt {kv_{k,i}}+\delta}}\leq \sum_{i=1}^{d}\frac{2(2-\gamma)}{\gamma}(\sqrt {tv_{t,i}}+\delta)  .$$
\end{Lemma}

{\bf Proof for Theorem \ref{adaptive individual bound}.}  Without loss of generality, we only prove Theorem \ref{adaptive individual bound} in the full gradient setting. It can be extended to stochastic cases using the regular technique in \citep{rakhlin2011making}.

Note that the projection operation can be rewritten as an optimization problem \citep{duchi2018introductory}, i.e., $\mathbf{w}_{t+1} = P_{\mathbf{Q}}[\mathbf{w}_{t} - \alpha_t \hat{V}_t^{-1}   \mathbf{g}(\mathbf{w}_t)+\beta_{1t}(\mathbf{w}_{t}-\mathbf{w}_{t-1})]$ is equivalent to
\begin{equation}\label{HB-regularization}
 \mathbf{w}_{t+1} = \arg\min_{\mathbf{w} \in \mathbf{Q}}\{ \alpha_t \hat{V}_t^{-1}\langle \mathbf{g}(\mathbf{w}_t),\mathbf{w}\rangle
 +\frac{1}{2} \|\mathbf{w}-\mathbf{w}_{t}-\beta_{1t}(\mathbf{w}_{t}-\mathbf{w}_{t-1})\|^2 \}.
\end{equation}
Then, $\forall\mathbf{u}\in \mathbf{Q}$, we have
$$
\langle\mathbf{w}_{t+1}-\mathbf{w}_{t}-\beta_t(\mathbf{w}_{t}-\mathbf{w}_{t-1})+\alpha_t \hat{V}_t^{-1}  \mathbf{g}(\mathbf{w}_t),\mathbf{w}_{t+1}-\mathbf{w}\rangle \leq 0.
$$
This is
\begin{equation}\label{proj8}
\langle\mathbf{w}_{t+1}+\mathbf{p}_{t+1}-(\mathbf{w}_{t}+\mathbf{p}_{t})+\frac{\alpha}{\sqrt t}\hat{V}_t^{-1}\mathbf{g}(\mathbf{w}_t),\mathbf{w}_{t+1}-\mathbf{w}\rangle \leq 0.
\end{equation}
Specifically,
\begin{equation}\label{proj9}
\langle\mathbf{w}_{t+1}+\mathbf{p}_{t+1}-(\mathbf{w}_{t}+\mathbf{p}_{t})+\frac{\alpha}{\sqrt t}\hat{V}_t^{-1}\mathbf{g}(\mathbf{w}_t),\mathbf{w}_{t+1}-\mathbf{w}_{t}\rangle \leq 0.
\end{equation}
From (\ref{proj8}) and (\ref{proj9}),
$$
\langle\mathbf{w}_{t+1}+\mathbf{p}_{t+1}-(\mathbf{w}_{t}+\mathbf{p}_{t})+\frac{\alpha}{\sqrt t}\hat{V}_t^{-1}\mathbf{g}(\mathbf{w}_t),\mathbf{w}_{t+1}-\mathbf{w}_t+(t+1)(\mathbf{w}_{t+1}-\mathbf{w}_t)\rangle \leq 0.
$$
i.e.,
$$
\langle\mathbf{w}_{t+1}+\mathbf{p}_{t+1}-(\mathbf{w}_{t}+\mathbf{p}_{t})+\frac{\alpha}{\sqrt t}\hat{V}_t^{-1}\mathbf{g}(\mathbf{w}_t),\mathbf{w}_{t+1}+\mathbf{p}_{t+1}-\mathbf{w}_t\rangle \leq 0.
$$
Using Lemma \ref{projection inequality}, we have
$$
\mathbf{w}_{t+1}+\mathbf{p}_{t+1}=P_{\mathbf{Q}}^{\hat{V}_t}[\mathbf{w}_{t}+\mathbf{p}_{t}-\frac{\alpha}{\sqrt t}\hat{V}_t^{-1}\mathbf{g}(\mathbf{w}_t)].
$$
Then
\begin{equation*}
\begin{aligned}
\|\mathbf{w}^*-(\mathbf{w}_{t+1}+\mathbf{p}_{t+1})\|_{\hat{V}_t}^2 
&\leq\|\mathbf{w}^*-(\mathbf{w}_{t}+\mathbf{p}_{t})+\frac{\alpha}{\sqrt t}\hat{V}_t^{-1}\mathbf{g}(\mathbf{w}_t)\|_{\hat{V}_t}^2\\
& =\|\mathbf{w}^*-(\mathbf{w}_{t}+\mathbf{p}_{t})\|_{\hat{V}_t}^2+\|\frac{\alpha}{\sqrt t}\mathbf{g}(\mathbf{w}_t)\|_{\hat{V}_t}^2 \\
&+2\langle\frac{\alpha}{\sqrt t}\mathbf{g}(\mathbf{w}_t),\mathbf{w}^*-\mathbf{w}_{t}\rangle+2\langle\frac{\alpha t}{\sqrt t}\mathbf{g}(\mathbf{w}_t),\mathbf{w}_{t-1}-\mathbf{w}_{t}\rangle.
\end{aligned}
\end{equation*}
Note
$$\langle\mathbf{g}(\mathbf{w}_t),\mathbf{w}^*-\mathbf{w}_{t}\rangle \leq f(\mathbf{w}^*)-f(\mathbf{w}_{t}),\ \langle\mathbf{g}(\mathbf{w}_t),\mathbf{w}_{t-1}-\mathbf{w}_{t}\rangle \leq f(\mathbf{w}_{t-1})-f(\mathbf{w}_{t}). $$
Then
$$
\begin{aligned}
(t+1)(f(\mathbf{w}_t)-f(\mathbf{w}^*)) 
\leq& t(f(\mathbf{w}_{t-1})-f(\mathbf{w}^*))+ \frac{\sqrt t}{2\alpha} \|\mathbf{w}^*-(\mathbf{w}_{t}+\mathbf{p}_{t})\|_{\hat{V}_t}^2\\
 -&\frac{\sqrt t}{2\alpha} \|\mathbf{w}^*-(\mathbf{w}_{t+1}+\mathbf{p}_{t+1})\|_{\hat{V}_t}^2 + \frac{\alpha}{2\sqrt t}\|\mathbf{g}(\mathbf{w}_t)\|_{\hat{V}_t^{-1}}^2.
\end{aligned}
$$
Summing this inequality from $k=1$ to $t$, we obtain
$$
\begin{aligned}
(t+1)(f(\mathbf{w}_t)-f(\mathbf{w}^*))
& \leq f(\mathbf{w}_{0})-f(\mathbf{w}^*)+ \sum_{k=1}^{t} {\frac{\alpha}{2\sqrt k}\|\mathbf{g}(\mathbf{w}_k)\|_{\hat{V}_k^{-1}}^2} \\
+ &\sum_{k=1}^{t}\big[ \frac{\sqrt k}{2\alpha} (\|\mathbf{w}^*-(\mathbf{w}_{k}+\mathbf{p}_{k})\|_{\hat{V}_k}^2  - \|\mathbf{w}^*-(\mathbf{w}_{k+1}+\mathbf{p}_{k+1})\|_{\hat{V}_k}^2) \big].
\end{aligned}
$$
Using Lemma \ref{square gradient}, we have
$$\sum_{k=1}^{t} {\frac{\alpha}{2\sqrt k}\|\mathbf{g}(\mathbf{w}_k)\|_{\hat{V}_k^{-1}}^2}\leq \sum_{i=1}^{d}\frac{\alpha(2-\gamma)}{\gamma}(\sqrt {tv_{t,i}}+\delta) . $$
Note
$$
\sqrt {v_{t,i}}\leq M.
$$
and
\begin{equation}\label{zinkevich1}
\begin{aligned}
&\sum_{k=1}^{t}\big[ \frac{\sqrt k}{2\alpha} (\|\mathbf{w}^*-(\mathbf{w}_{k}+\mathbf{p}_{k})\|_{\hat{V}_k}^2  - \|\mathbf{w}^*-(\mathbf{w}_{k+1}+\mathbf{p}_{k+1})\|_{\hat{V}_k}^2) \big]\\
& =\sum_{i=1}^{d}\frac{\hat{V}_1}{2\alpha} \|\mathbf{w}^{*}-(\mathbf{w}_{1}+\mathbf{p}_{1})\|^2- \sum_{i=1}^{d}\frac{\sqrt t\hat{V}_t}{2\alpha} \|\mathbf{w}^*-(\mathbf{w}_{t+1}+\mathbf{p}_{t+1})\|^2  \\
&+\sum_{i=1}^{d}\sum_{k=2}^{t}\frac{1}{2\alpha}(\sqrt k \hat{v}_{k,i}-\sqrt{k-1}\hat{v}_{k-1,i}) \|\mathbf{w}^*-(\mathbf{w}_{k}+\mathbf{p}_{k})\|^2.
\end{aligned}
\end{equation}
Since $\mathbf{Q}$ is a bounded set, there exists a positive number $M_0 >0$ such that
$$
\begin{aligned}
\|\mathbf{w}^*-(\mathbf{w}_{t+1}+\mathbf{p}_{t+1})\|^2 \leq M_0, \forall t \geq 0.
\end{aligned}
$$
and $v_{k,i}=\beta_{2k}{v}_{k-1,i}+(1-\beta_{2k})g_{k,i}^2$ as well as $\beta_{2k}\geq 1-\frac{1}{k}$ which implies $k\beta_{2k}\geq k-1$, we get
$$
\begin{aligned}
\sqrt k \hat{v}_{k,i}&=\sqrt {k {v}_{k,i}}+\delta\\
&=\sqrt {k \beta_{2k}{v}_{k-1,i}+k(1-\beta_{2k})g_{k,i}^2}+\delta\\
&\geq\sqrt{(k-1)v_{k-1,i}}+\delta\\
&=\sqrt{k-1}\hat{v}_{k-1,i}.
\end{aligned}
$$
From (\ref{zinkevich1}) we have
\begin{equation}\label{zinkevich2}
\begin{aligned}
&\sum_{k=1}^{t}\big[ \frac{\sqrt k}{2\alpha} (\|\mathbf{w}^*-(\mathbf{w}_{k}+\mathbf{p}_{k})\|_{\hat{V}_k}^2  - \|\mathbf{w}^*-(\mathbf{w}_{k+1}+\mathbf{p}_{k+1})\|_{\hat{V}_k}^2) \big]\\
\leq&\sum_{i=1}^{d}\frac{\hat{v}_{1,i}}{2\alpha} M_{0}+\sum_{i=1}^{d}\sum_{k=2}^{t}\frac{1}{2\alpha}(\sqrt k \hat{v}_{k,i}-\sqrt{k-1}\hat{v}_{k-1,i}) M_{0}\\
=&\frac{d\hat{v}_{1,i} M_{0}}{2\alpha} +\frac{d\sqrt t \hat{v}_{t,i}M_{0}}{2\alpha} -\frac{d\hat{v}_{1,i}M_{0}}{2\alpha}\\
\leq &\frac{d(\sqrt t M+\delta)M_{0}}{2\alpha}.
\end{aligned}
\end{equation}
Therefore
$$
\begin{aligned}
& (t+1)[f(\mathbf{w}_t)-f(\mathbf{w}^*)] \leq f(\mathbf{w}_{0})-f(\mathbf{w}^*)+ \frac{d\alpha(2-\gamma)(\sqrt tM+\delta) }{\gamma} +\frac{d(\sqrt t M+\delta)M_{0}}{2\alpha} .
\end{aligned}
$$

This proves
$$
f(\mathbf{w}_{t})-f(\mathbf{w}^*) \leq O( \frac{1}{\sqrt{t}}).
$$

\subsection{Proof for Theorem \ref{adaptive averaging bound}}

 Let $\{{\mathbf{w}}_{t}\}_{t=1}^{\infty}$ be generated by the adaptive HB (Algorithm \ref{alg.hb}). Set
$$\mathbf{p}_t=\frac{\beta}{1-\beta}(\mathbf{w}_{t}-\mathbf{w}_{t-1}) \ \text{and} \  \alpha_t=\displaystyle\frac{\alpha}{\sqrt{t}}.$$
Then, $\forall\mathbf{u}\in \mathbf{Q}$, according to Lemma \ref{projection inequality}, we have
$$
\langle\mathbf{w}_{t+1}-\mathbf{w}_{t}-\beta(\mathbf{w}_{t}-\mathbf{w}_{t-1})+ \alpha_t \hat{V}_t^{-1} \mathbf{g}(\mathbf{w}_t),\mathbf{w}_{t+1}-\mathbf{w}\rangle \leq 0.
$$
This is
$$
\langle\frac{1}{1-\beta}(\mathbf{w}_{t+1}-\mathbf{w}_{t})-\mathbf{p}_{t}+\frac{\alpha_t \hat{V}_t^{-1}}{1-\beta}\mathbf{g}(\mathbf{w}_t),
\mathbf{w}_{t+1}-\mathbf{w}\rangle \leq 0.
$$
i.e.,
\begin{equation}\label{proj10}
\langle\mathbf{w}_{t+1}+\mathbf{p}_{t+1}-(\mathbf{w}_{t}+\mathbf{p}_{t})+\frac{\alpha_t \hat{V}_t^{-1}}{1-\beta}\mathbf{g}(\mathbf{w}_t),\mathbf{w}_{t+1}-\mathbf{w}\rangle \leq 0
\end{equation}
Specifically,
\begin{equation}\label{proj11}
\langle\mathbf{w}_{t+1}+\mathbf{p}_{t+1}-(\mathbf{w}_{t}+\mathbf{p}_{t})+\frac{\alpha_t \hat{V}_t^{-1}}{1-\beta}\mathbf{g}(\mathbf{w}_t),
\frac{\beta(\mathbf{w}_{t+1}-\mathbf{w}_t)}{1-\beta}\rangle \leq 0
\end{equation}
From (\ref{proj10}) and (\ref{proj11}),
$$
\langle\mathbf{w}_{t+1}+\mathbf{p}_{t+1}-(\mathbf{w}_{t}+\mathbf{p}_{t})+\frac{\alpha_t \hat{V}_t^{-1}}{1-\beta}\mathbf{g}(\mathbf{w}_t),
\mathbf{w}_{t+1}+\mathbf{p}_{t+1}-\mathbf{w}\rangle \leq 0.
$$
Using Lemma \ref{projection inequality}, we have
$$
\mathbf{w}_{t+1}+\mathbf{p}_{t+1}=P_{\mathbf{Q}}^{\hat{V}_t}[\mathbf{w}_{t}+\mathbf{p}_{t}-\displaystyle\frac{\alpha_t \hat{V}_t^{-1}}{1-\beta}\mathbf{g}(\mathbf{w}_{t})].
$$

According to Lemma \ref{constrained and unconstrained},
\begin{equation*}
\begin{split}
&\|\mathbf{w}^*-(\mathbf{w}_{t+1}+\mathbf{p}_{t+1})\|_{\hat{V}_t}^2 \\
\leq  & \|\mathbf{w}^*-(\mathbf{w}_{t}+\mathbf{p}_{t})+\frac{\alpha_t \hat{V}_t^{-1}}{1-\beta}\mathbf{g}(\mathbf{w}_t)\|_{\hat{V}_t}^2\\
=&\|\mathbf{w}^*-(\mathbf{w}_{t}+\mathbf{p}_{t})\|_{\hat{V}_t}^2+\|\frac{\alpha_t}{1-\beta}\mathbf{g}(\mathbf{w}_t)\|_{\hat{V}_t}^2\\
+&2\langle\frac{\alpha_t}{1-\beta}\mathbf{g}(\mathbf{w}_t),\mathbf{w}^*-\mathbf{w}_{t}\rangle+2\langle\frac{\alpha_t\beta }{(1-\beta)^2}\mathbf{g}(\mathbf{w}_t),\mathbf{w}_{t-1}-\mathbf{w}_{t}\rangle
\end{split}
\end{equation*}
Note
$$\langle\mathbf{g}(\mathbf{w}_t),\mathbf{w}^*-\mathbf{w}_{t}\rangle \leq f(\mathbf{w}^*)-f(\mathbf{w}_{t}),\ \langle\mathbf{g}(\mathbf{w}_t),\mathbf{w}_{t-1}-\mathbf{w}_{t}\rangle \leq f(\mathbf{w}_{t-1})-f(\mathbf{w}_{t}). $$
Then
$$
\begin{aligned}
& \|\mathbf{w}^*-(\mathbf{w}_{t+1}+\mathbf{p}_{t+1})\|_{ \hat{V}_t}^2 \\
\leq& \|\mathbf{w}^*-(\mathbf{w}_{t}+\mathbf{p}_{t})\|_{\hat{V}_t}^2+
\frac{\alpha_t^2}{(1-\beta)^2}\|\mathbf{g}(\mathbf{w}_t)\|_{\hat{V}_t^{-1}}^2\\
+&\frac{2\alpha_t}{1-\beta}[f(\mathbf{w}^*)-f(\mathbf{w}_{t})]+
\frac{2\alpha_t\beta}{(1-\beta)^2}[f(\mathbf{w}_{t-1})-f(\mathbf{w}_{t})].
\end{aligned}
$$
Rearrange the inequality, we have
$$
\begin{aligned}
&\frac{2\alpha_t}{1-\beta}[f(\mathbf{w}_t)-f(\mathbf{w}^*)]\\
\leq&\frac{2\alpha_t\beta}{(1-\beta)^2}[f(\mathbf{w}_{t-1})-f(\mathbf{w}_t)]+
\|\mathbf{w}^*-(\mathbf{w}_t+\mathbf{p}_t)\|_{\hat{V}_t}^2\\
-&\|\mathbf{w}^*-(\mathbf{w}_{t+1}+\mathbf{p}_{t+1})\|_{\hat{V}_t}^2
+\frac{\alpha_t^2}{(1-\beta)^2}\|\mathbf{g}(\mathbf{w}_t)\|_{\hat{V}_t^{-1}}^2.
\end{aligned}
$$
i.e.,
$$
\begin{aligned}
&f(\mathbf{w}_t)-f(\mathbf{w}^*)\\
\leq&\frac{\beta}{1-\beta}[f(\mathbf{w}_{t-1})-f(\mathbf{w}_t)]+
\frac{1-\beta}{2\alpha_t}[\|\mathbf{w}^*-(\mathbf{w}_t+\mathbf{p}_t)\|_{\hat{V}_t}^2\\
-&\|\mathbf{w}^*-(\mathbf{w}_{t+1}+\mathbf{p}_{t+1})\|_{\hat{V}_t}^2]+
\frac{\alpha_t}{2(1-\beta)}\|\mathbf{g}(\mathbf{w}_t)\|_{\hat{V}_t^{-1}}^2.
\end{aligned}
$$
Summing this inequality from $k=1$ to $t$, we obtain
$$
\begin{aligned}
& \sum_{k=1}^{t}[{f(\mathbf{w}_k)-f(\mathbf{w}^*)}]\\
\leq&\frac{\beta}{1-\beta}[f(\mathbf{w}_{0})-f(\mathbf{w}_t)]+
\frac{1-\beta}{2\alpha_1}\|\mathbf{w}^*-(\mathbf{w}_{1}+\mathbf{p}_{1})\|_{\hat{V}_1}^2\\
-&\frac{1-\beta}{2\alpha_t}\|\mathbf{w}^*-(\mathbf{w}_{t+1}+\mathbf{p}_{t+1})\|_{\hat{V}_t}^2
+\sum_{k=1}^{t} {\frac{\alpha_k}{2(1-\beta)}\|\mathbf{g}(\mathbf{w}_k)\|_{\hat{V}_k^{-1}}^2}\\
+&\sum_{i=1}^{d}\sum_{k=2}^{t}{\|\mathbf{w}^*-(\mathbf{w}_{k}+\mathbf{p}_{k})\|^2(\frac{(1-\beta)\hat{v}_{k,i}}{2\alpha_k}-\frac{(1-\beta)\hat{v}_{k-1,i}}{2\alpha_{k-1}})}.
\end{aligned}
$$
i.e.,
\begin{equation}\label{proj12}
\begin{aligned}
& \sum_{k=1}^{t}[{f(\mathbf{w}_k)-f(\mathbf{w}^*)}]\\
\leq&\frac{\beta}{1-\beta}[f(\mathbf{w}_{0})-f(\mathbf{w}_t)]+
\frac{1-\beta}{2\alpha}\|\mathbf{w}^*-(\mathbf{w}_{1}+\mathbf{p}_{1})\|_{\hat{V}_1}^2\\
+&\sum_{i=1}^{d}\sum_{k=2}^{t}{\|\mathbf{w}^*-(\mathbf{w}_{k}+\mathbf{p}_{k})\|^2\frac{1-\beta}{2\alpha}(\sqrt{k}\hat{v}_{k,i}-\sqrt{k-1}\hat{v}_{k-1,i})}\\
+&\sum_{k=1}^{t} {\frac{\alpha}{2(1-\beta)\sqrt{k}}\|\mathbf{g}(\mathbf{w}_k)\|_{\hat{V}_k^{-1}}^2}.
\end{aligned}
\end{equation}
Using Lemma \ref{square gradient}, we have
\begin{equation}\label{proj13}
\sum_{k=1}^{t} {\frac{\alpha}{2\sqrt k(1-\beta)}\|\mathbf{g}(\mathbf{w}_k)\|_{\hat{V}_k^{-1}}^2}\leq \sum_{i=1}^{d}\frac{\alpha(2-\gamma)}{\gamma(1-\beta)}(\sqrt {tv_{t,i}}+\delta)\leq\frac{da(2-\gamma)(\sqrt{t}M+\delta)}{\gamma(1-\beta)} .
\end{equation}
and since $\mathbf{Q}$ is a bounded set, there exists a positive number $M_0 >0$ such that
\begin{equation}\label{proj14}
\|\mathbf{w}^*-(\mathbf{w}_{t+1}+\mathbf{p}_{t+1})\|^2 \leq M_0, \forall t \geq 0.
\end{equation}
From (\ref{proj12})(\ref{proj13})(\ref{proj14}) we have,
$$
\begin{aligned}
&\sum_{k=1}^{t}[{f(\mathbf{w}_k)-f(\mathbf{w}^*)}]\\
\leq&\frac{\beta}{1-\beta}[f(\mathbf{w}_{0})-f(\mathbf{w}_t)]+
\sum_{i=1}^{d}\frac{(1-\beta)\hat{v}_{1,i}M_0}{2\alpha}+\frac{da(2-\gamma)(\sqrt{t}M+\delta)}{\gamma(1-\beta)}\\
+&\sum_{i=1}^{d}\frac{(1-\beta)\hat{v}_{t,i}\sqrt{t}M_0}{2\alpha}-\sum_{i=1}^{d}\frac{(1-\beta)\hat{v}_{1,i}M_0}{2\alpha}.
\end{aligned}
$$
i.e.,
$$
\sum_{k=1}^{t}[{f(\mathbf{w}_k)-f(\mathbf{w}^*)}]
\leq\frac{\beta}{1-\beta}[f(\mathbf{w}_{0})-f(\mathbf{w}_t)]+
\frac{da(2-\gamma)(\sqrt{t}M+\delta)}{\gamma(1-\beta)}+
\frac{d(1-\beta)(\sqrt tM+\delta)M_0}{2\alpha}.
$$
By convexity of $f(\mathbf{w}_k)$, we obtain
$$
f(\frac{1}{t}\sum_{k=1}^{t}\mathbf{w}_{k})-f(\mathbf{w}^*)
\leq\frac{\beta}{(1-\beta)t}[f(\mathbf{w}_{0})-f(\mathbf{w}_t)]+
\frac{da(2-\gamma)(\sqrt{t}M+\delta)}{\gamma(1-\beta)t}+
\frac{d(1-\beta)(\sqrt tM+\delta)M_0}{2\alpha t}.
$$

This completes the proof of Theorem \ref{adaptive averaging bound}.

\subsection{Experiments on Optimizing a Synthetic Convex Function}

A constrained convex optimization problem  was constructed in \citep{harvey2019tight}. Let $\mathbf{Q}$ be unit ball in $\mathbb{R}^T$. For $ i\in[T]$ and $c\ge1$, define the positive scalar parameters
\begin{equation*}
  a_i=\frac{1}{8c(T-i+1)}\qquad
  b_i=\frac{\sqrt{i}}{2c\sqrt{T}}
\end{equation*}

Define $f: \mathbf{Q}\to\mathbb{R}$ and $h_i\in\mathbb{R}^T $ for $i\in[T+1]$ by
\begin{equation*}
f(x)=\max \limits_{i\in[T]} \mathbf{h}_i^\top \mathbf{w}\qquad where \qquad
h_{i,j}=\left\{
    \begin{aligned}
    a_j & , &1\leq j<i \\
    -b_j & , & i=j<T \\
    0 & , & i<j\leq T
    \end{aligned}
            \right.
\end{equation*}

\begin{figure}[ht]
    \centering
    \includegraphics[width=0.8\textwidth]{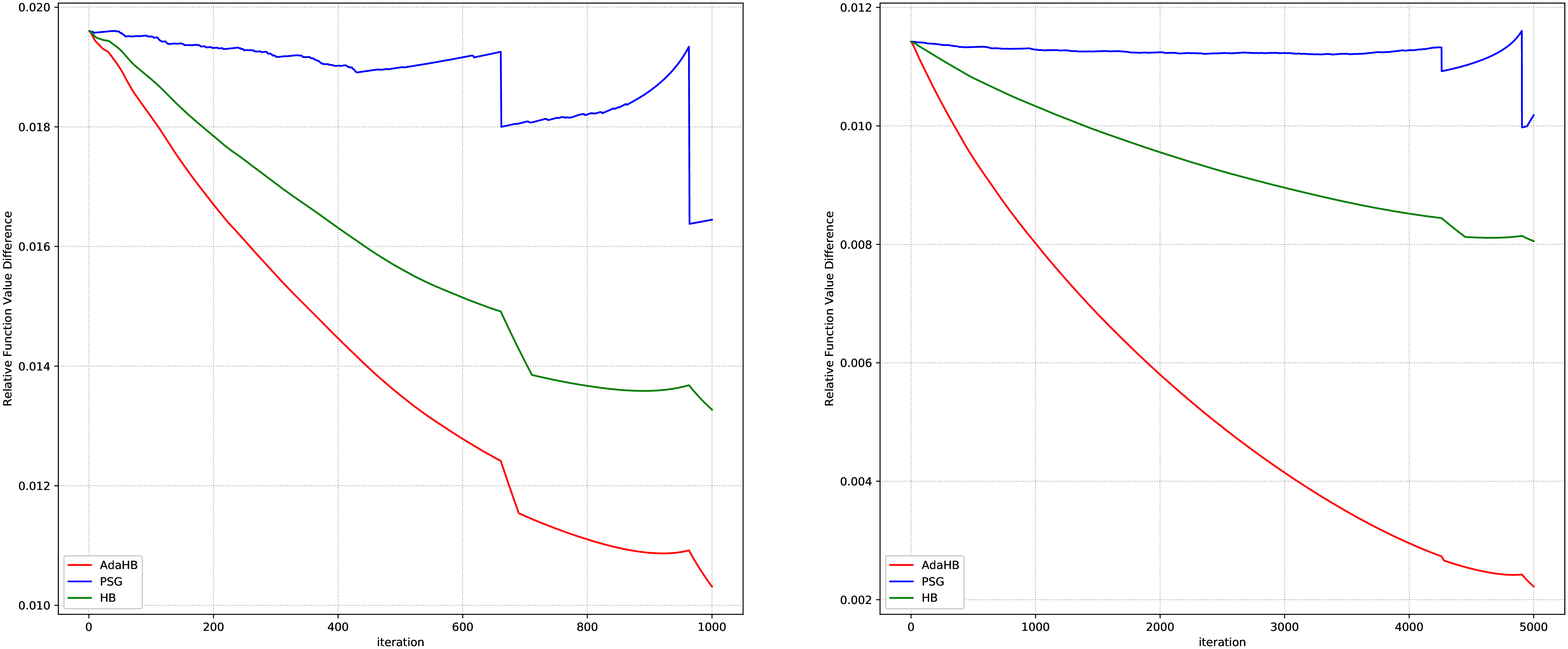}
    \caption{Convergence of the function value when $T=1000$ and $T=5000$}
    \label{fig:artificial experiment}
\end{figure}

Obviously, the minimum value of $f$ on the unit ball is non-positive because $f(0)=0$. It can be proved $f(\mathbf{w}_T) \geq \frac{\log T}{32c\sqrt{T}}$. Set $c=2$, the function value $f(\mathbf{w}_t) $ v.s. iteration is illustrated in Figure \ref{fig:artificial experiment}, where the stepsize of GD is $\frac{c}{\sqrt{t}}$ and the parameters of the constrained HB (\ref{constrained HB}) ($\alpha=8$) and AdaHB (\ref{adaptive HB}) ($\alpha=0.08$, $\gamma=0.9$, $\delta=10^{-8}$) are selected according to Theorem \ref{individual bound} and Theorem \ref{adaptive individual bound}. As expected, the individual convergence of HB is much faster than that of PSG. We thus conclude that HB is an effective acceleration of GD in terms of the individual convergence.

\subsection{Experiments on Optimizing General Convex Functions}

We consider the hinge loss optimization problem with $l_1$-ball constraints and use SLEP package\footnote{\url{http://yelabs.net/software/SLEP/}} for $l_1$ projection operation.
\begin{equation}\label{l1-regularization minimization}
\min f(\mathbf{w}), \ s. t. \  \mathbf{w} \in \{\mathbf{w}: {\|\mathbf{w}\|}_1 \leq \tau \}.
\end{equation}
Datasets: A9a, W8a, Covtype, Ijcnn1, Rcv1, Realsim (available at LibSVM\footnote{\url{http://www.csie.ntu.edu.tw/~cjlin/libsvmtools/datasets/}} website).

Algorithms: PSG ($\alpha_t = \frac{\alpha}{\sqrt t}$), HB ($\alpha_t = \frac{\alpha}{(t+2)\sqrt{t}}$, $\beta_t= \frac{t}{t+2}$), NAG \citep{srsg2019} and adaptive HB (\ref{adaptive HB}) ($\beta_{1t}=\frac{t}{t+2}$).

\begin{figure}[ht]
    \centering
    \subfigure[Covtype ($\tau=50$)]{
    \includegraphics[width=2.5in]{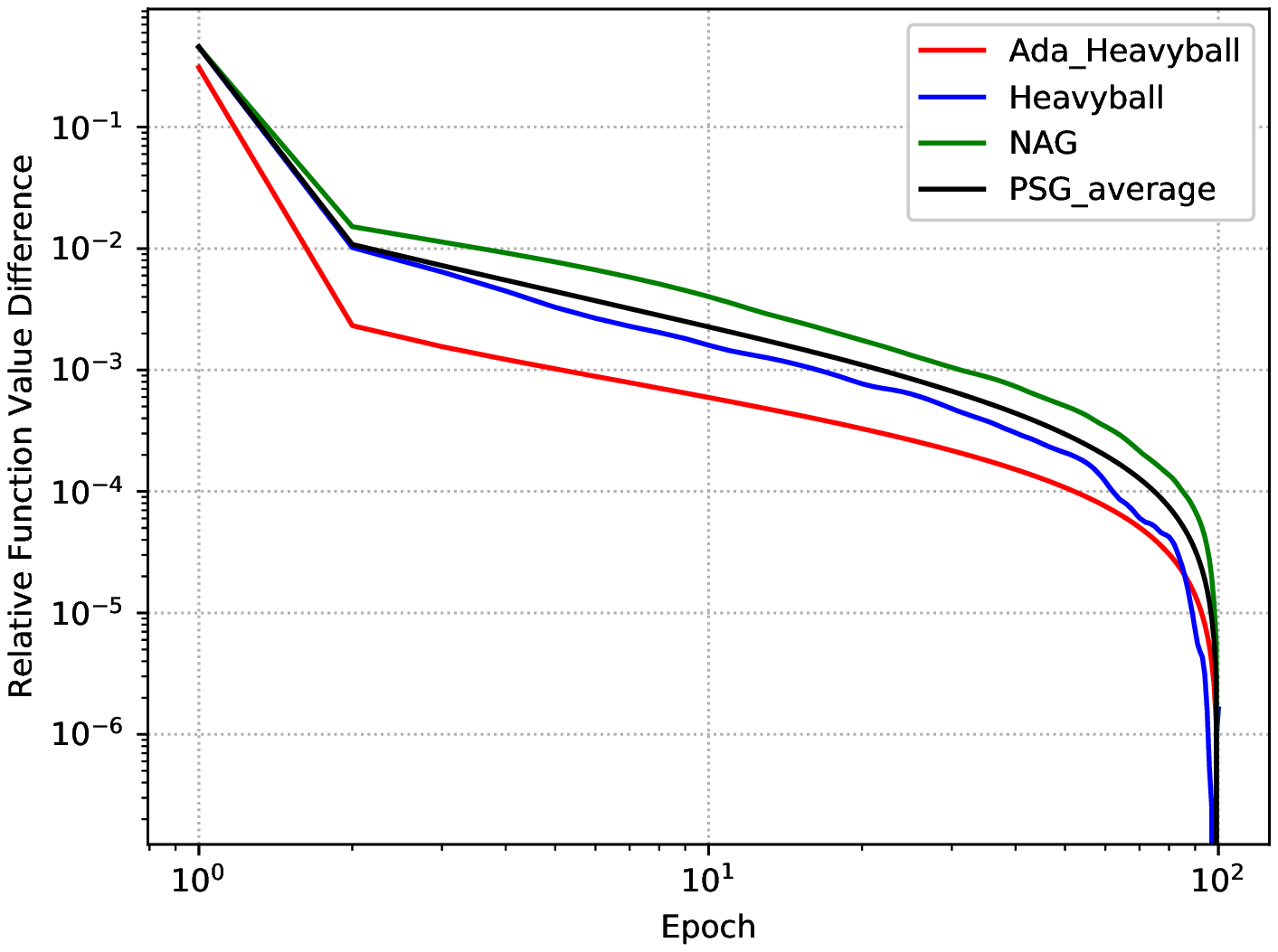}}
    \subfigure[Realsim ($\tau=60$)]{
    \includegraphics[width=2.5in]{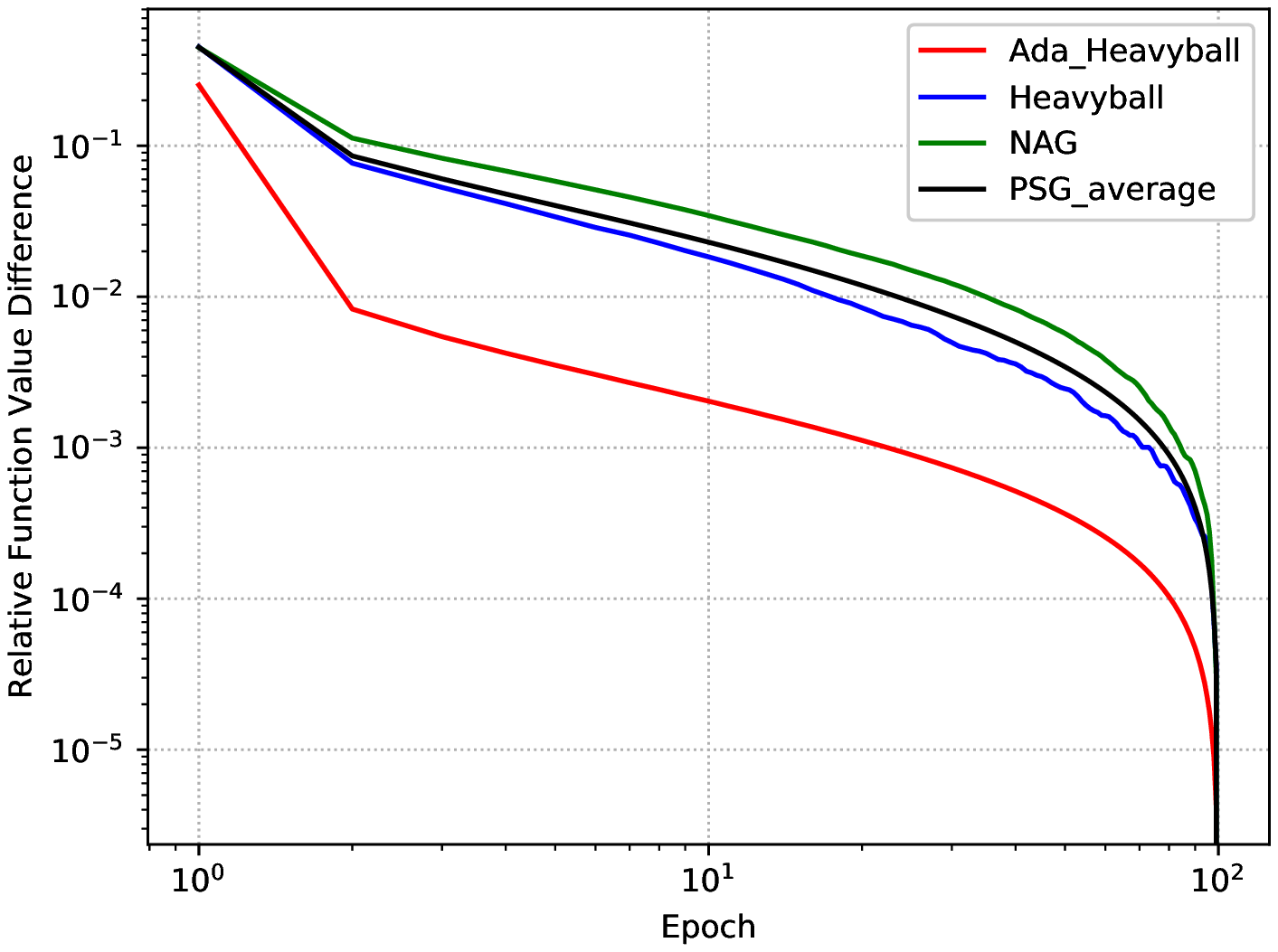}}
    \subfigure[A9a ($\tau=20$)]{
    \includegraphics[width=2.5in]{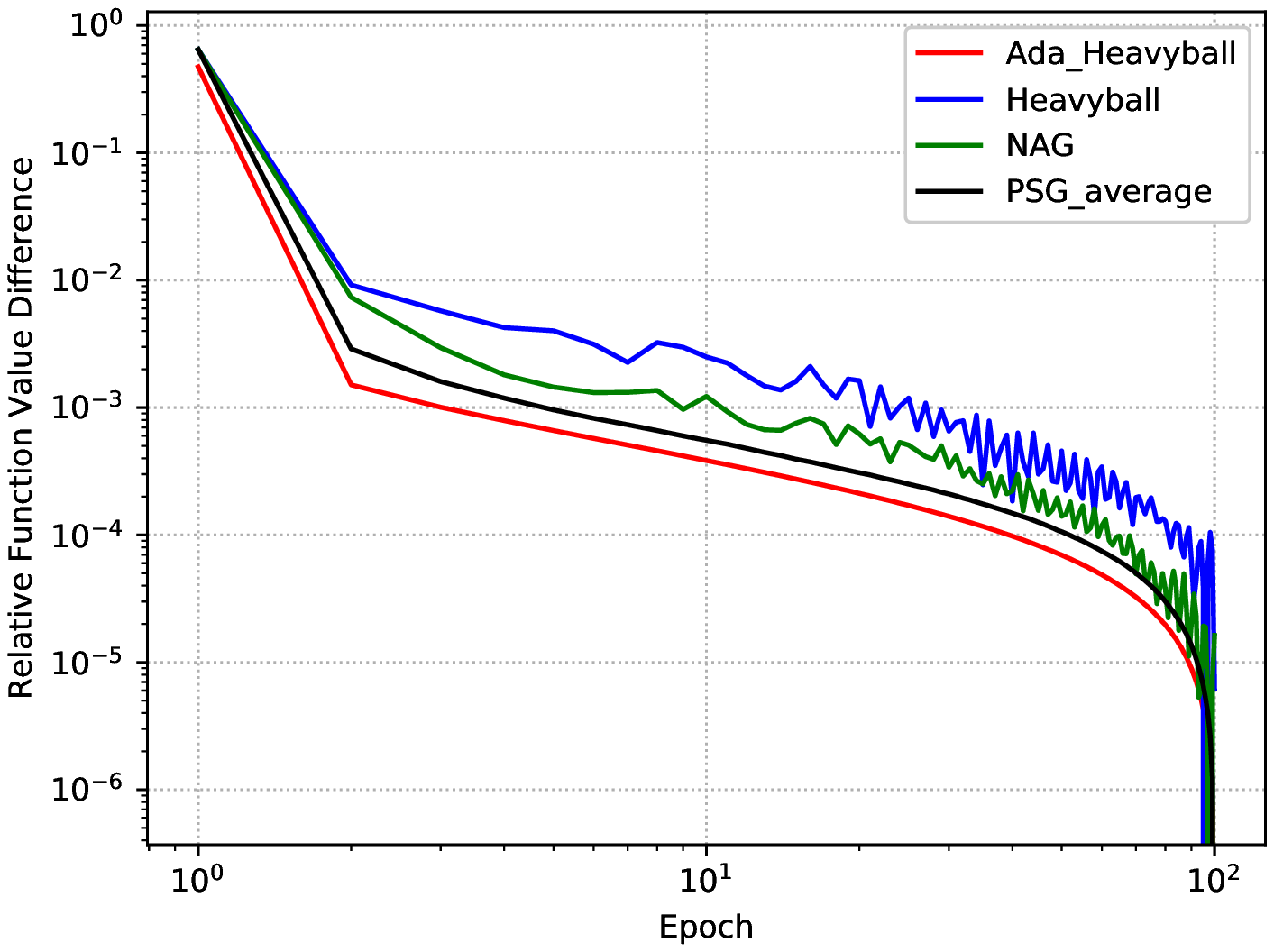}}
    \subfigure[W8a ($\tau=30$)]{
    \includegraphics[width=2.5in]{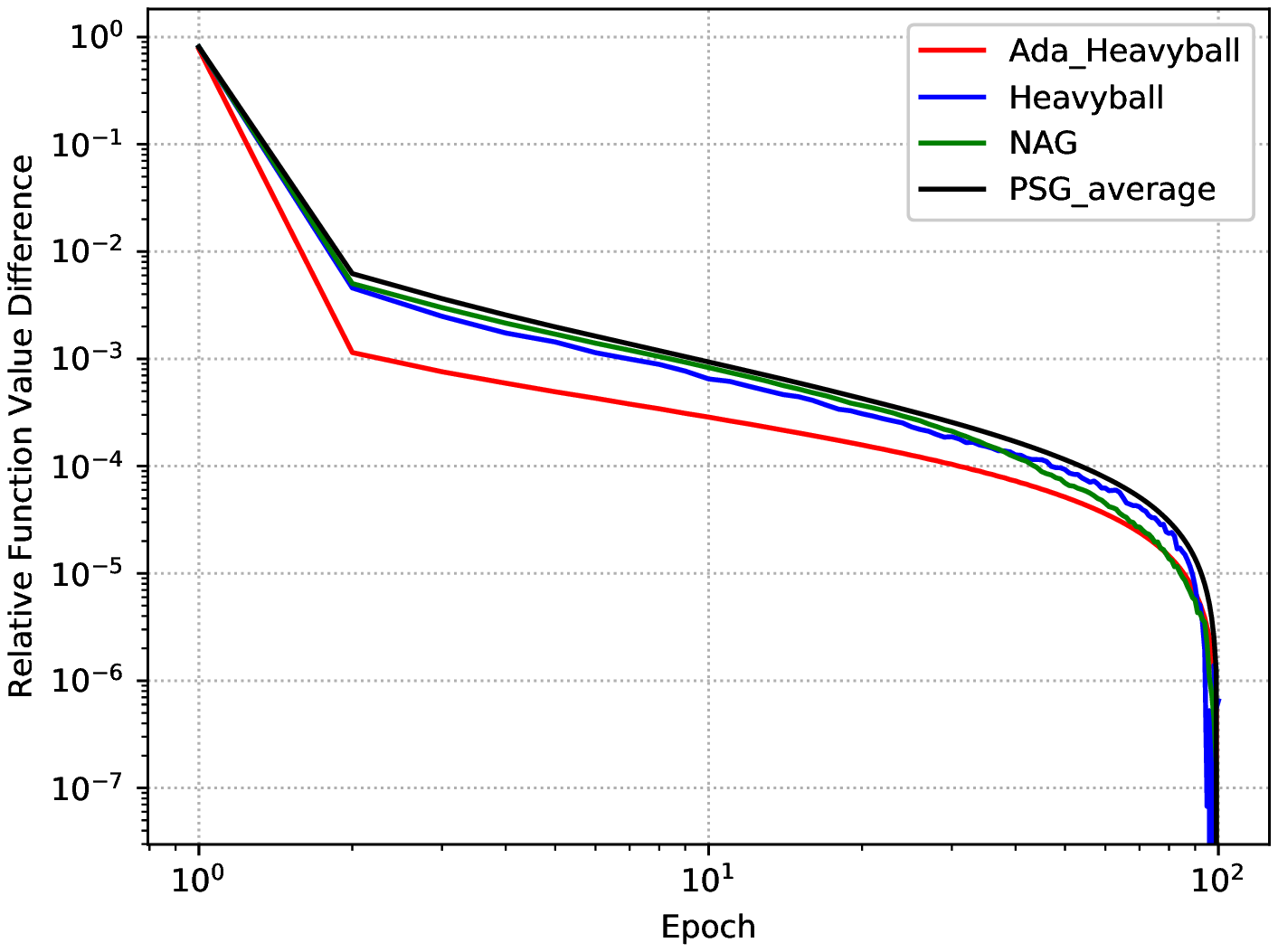}}
    \subfigure[Ijcnn1 ($\tau=10$)]{
    \includegraphics[width=2.5in]{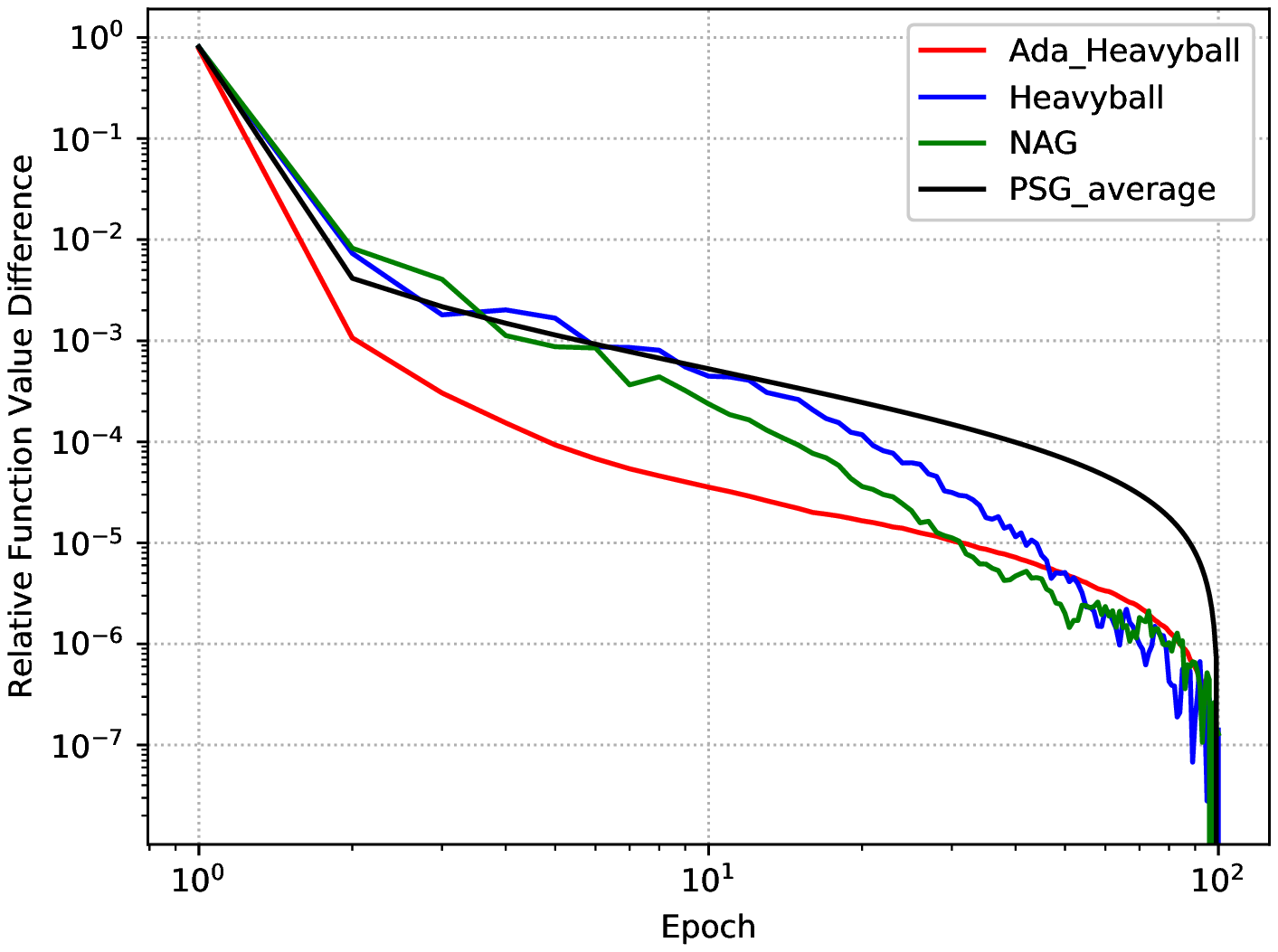}}
    \subfigure[Rcv1 ($\tau=80$)]{
    \includegraphics[width=2.5in]{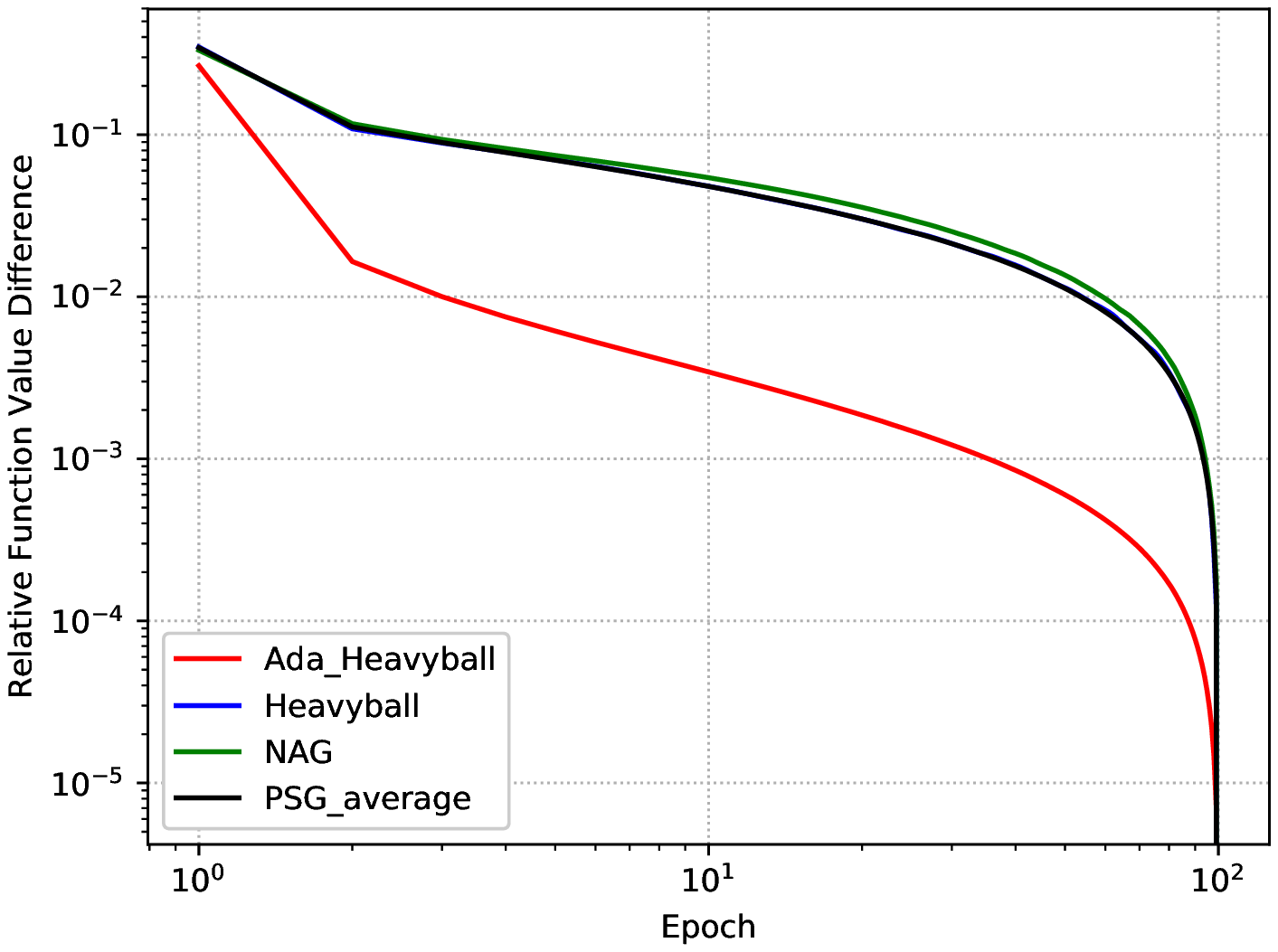}}
    \caption{Convergence on different LibSVM datasets for $l_1$ hinge loss problems}
    \label{fig:convex experiment}
\end{figure}

The relative function value $f(\mathbf{w}_t)-f(\mathbf{w_*})$ v.s. epoch is illustrated in Figure \ref{fig:convex experiment}. As expected, the individual convergence of the adaptive HB has almost the same behavior as the averaging output of PSG, and the individual output of HB and NAG. Since the three stochastic methods have the optimal convergence, we conclude that the stochastic adaptive HB attains the optimal individual convergence for general convex regularized learning problems.

\end{document}